\documentclass[10pt,twocolumn,letterpaper]{article}

\usepackage{cvpr}              % To produce the CAMERA-READY version
\usepackage{indentfirst}
\usepackage{comment}
\usepackage{amssymb,pifont}
\newcommand{\cmark}{\ding{51}}%
\newcommand{\xmark}{\ding{55}}%
\usepackage{times}
\usepackage{epsfig}
\usepackage{graphicx}
\usepackage{amsmath}
\usepackage{array}
\usepackage{subcaption}
\usepackage{multirow}
\usepackage{subcaption}
\usepackage{indentfirst}
\usepackage{verbatim}
\usepackage{bm}
\usepackage{kantlipsum}
\usepackage{hhline}
\usepackage{color}
\usepackage{booktabs}
\usepackage{bbding}
\usepackage{makecell} 
\usepackage{colortbl} % Add this to your preamble for table coloring
\usepackage{xcolor}
\usepackage{amssymb}
\usepackage[table]{xcolor}
\definecolor{lightblue}{RGB}{220,240,255}
\definecolor{lightpurple}{RGB}{230,220,255}
\definecolor{lightgreen}{RGB}{220,255,220}
\definecolor{cvprblue}{rgb}{0.21,0.49,0.74}
\usepackage[pagebackref,breaklinks,colorlinks,allcolors=cvprblue]{hyperref}
\usepackage{array, tabularx, makecell}
\usepackage{ragged2e} % <-- 텍스트 줄바꿈 제어용
\definecolor{pastellavender}{RGB}{230,220,250}
\definecolor{pastelcream}{RGB}{255,249,214}

\def\paperID{40762} % *** Enter the Paper ID here
\def\confName{CVPR}
\def\confYear{2026}

\title{Stake the Points: Structure-Faithful Instance Unlearning}

\author{Kiseong Hong\\
Chung-Ang University\\
{\tt\small ghdrltjd@cau.ac.kr}
\and
JungKyoo Shin\\
Chung-Ang University\\
{\tt\small neo293@cau.ac.kr}
\and
Eunwoo Kim$\thanks{Corresponding author.}$\\
Chung-Ang University\\
\tt\small eunwoo@cau.ac.kr
}

\begin{document}
\maketitle

%%%%%%%%% ABSTRACT
\begin{abstract}
Machine unlearning (MU) addresses privacy risks in pretrained models. 
The main goal of MU is to remove the influence of designated data while preserving the utility of retained knowledge.
Achieving this goal requires preserving semantic relations among retained instances, which existing studies often overlook.
We observe that without such preservation, models suffer from progressive structural collapse, undermining both the deletion–retention balance.
In this work, we propose a novel structure-faithful framework that introduces stakes, i.e., semantic anchors that serve as reference points to maintain the knowledge structure.
By leveraging these anchors, our framework captures and stabilizes the semantic organization of knowledge.
Specifically, we instantiate the anchors from language-driven attribute descriptions encoded by a semantic encoder (e.g., CLIP).
We enforce preservation of the knowledge structure via structure-aware alignment and regularization: the former aligns the organization of retained knowledge before and after unlearning around anchors, while the latter regulates updates to structure-critical parameters.
Results from image classification, retrieval, and face recognition show average gains of 32.9\%, 22.5\%, and 19.3\% in performance, balancing the deletion–retention trade-off and enhancing generalization.
\end{abstract}

%%%%%%%%% BODY TEXT
\section{Introduction}
Requests to erase sensitive or personally related learned information are rising due to stronger data protection regulations \cite{voigt2017eu}.
A straightforward solution is to retrain from scratch without the designated data \cite{bourtoule2021machine}. 
However, this is often impractical due to limited data access and the high cost of large-scale retraining \cite{bourtoule2021machine, cao2015towards}. 
These challenges fuel the rapid growth of machine unlearning (MU), which aims to selectively remove the influence of designated data while preserving the utility of retained knowledge \cite{koh2017understanding, golatkar2020eternal, thudi2022unrolling, spartalis2025lotus}.

\begin{figure}[t]
\begin{center}
\includegraphics[width=\columnwidth]{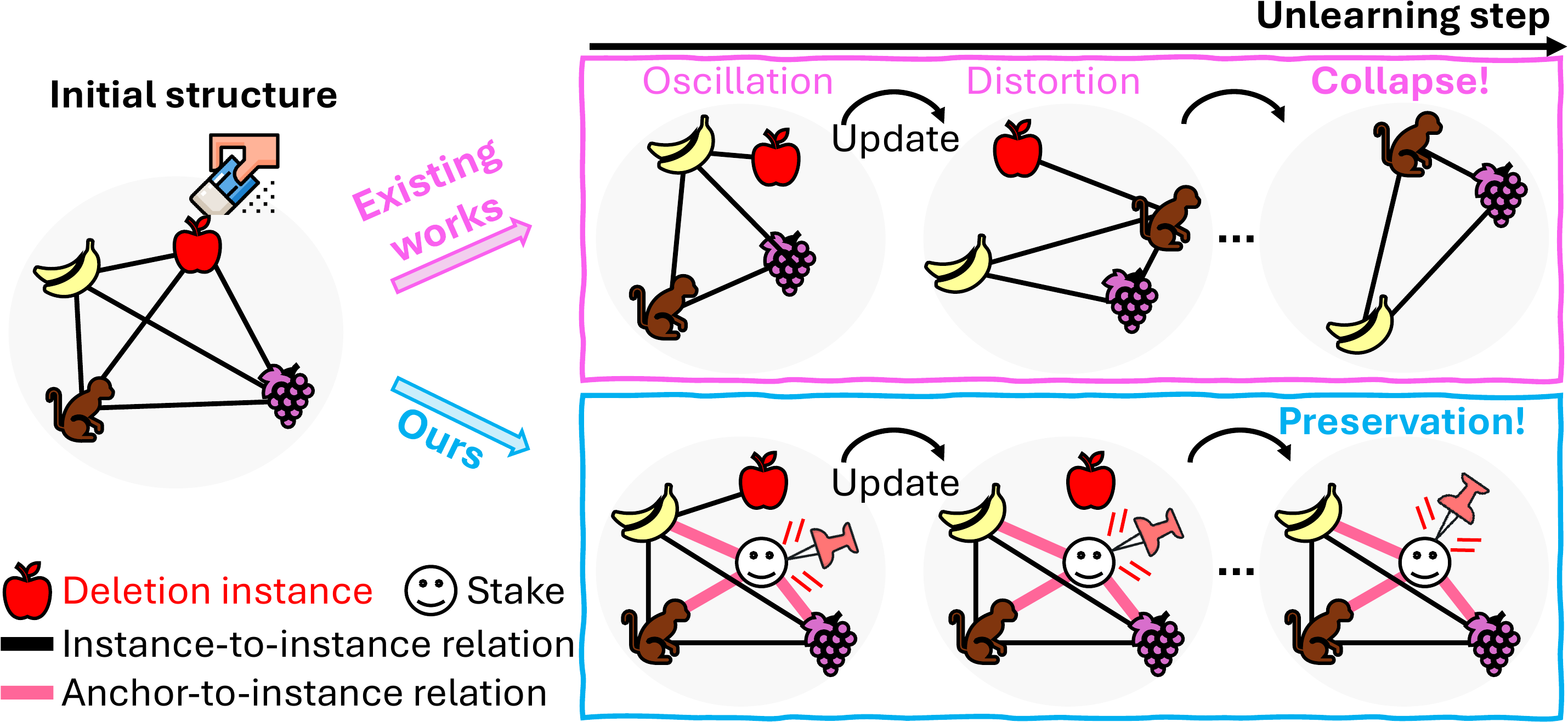}
\vspace{-9.mm}
\end{center}
\caption{
Conceptual illustration of structural collapse and our structure-faithful remedy in MU.
Existing works delete the designated instance but ignore semantic relations between retained instances.
During unlearning, model updates induce oscillations in the representation space.
Without relational awareness, these oscillations distort the instance-level semantic relations (e.g., a monkey embedding drifting toward grape while moving away from banana), collapsing the original knowledge organization.
We introduce semantic anchors (i.e., stakes) that preserve key relational constraints by keeping the relative positions between anchors and retained instances (i.e., anchor-to-instance relations).}
\label{fig:1}
\end{figure}

Research on MU advances from exact retraining approaches \cite{bourtoule2021machine, graves2021amnesiac, wang2022federated} to more efficient approximate methods \cite{lee2025esc, kim2024negmerge, cha2024learning, spartalis2025lotus}.
To improve efficiency, these approximate methods update parameters affected by the forget set~\cite{foster2024fast}, leverage weight saliency~\cite{fan2023salun}, or apply selective perturbations~\cite{golatkar2020eternal}, avoiding full retraining.
Recently, growing attention has been devoted toward instance-level unlearning \cite{cha2024learning, spartalis2025lotus}, as real-world deletion requests typically involve specific individuals rather than entire categories \cite{mehrabi2021survey, unlearning_challenge2023}.

Despite these advances, the core challenge of MU lies in preserving structure, particularly the semantic organization among the retained instances.
This is because knowledge in deep models is encoded not in isolated knowledge but in the semantic relations that contextualize them \cite{bollacker2008freebase, zhang2020connecting, hong2025rainbowprompt}.
However, existing methods \cite{cha2024learning, golatkar2020eternal} largely neglect this aspect.
As shown in Figure~\ref{fig:1}, removing an instance (e.g., apple) can distort the semantic relations between other instances (e.g., banana or grape).
Such distortion through the representation space disrupts coherence, leading to progressive structural collapse.
This highlights the necessity of structure-preserving unlearning to maintain relational structure.

\begin{figure}[t]
\centering
\includegraphics[width=\columnwidth]{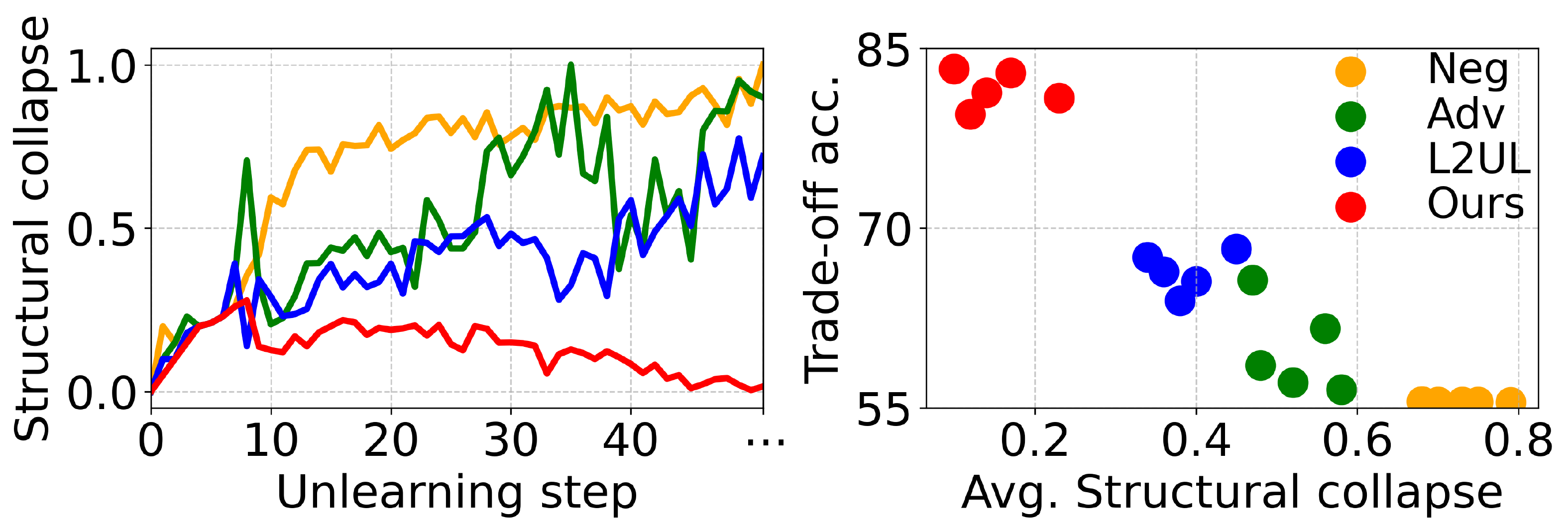}
\vspace{-1.mm}
\begin{minipage}[t]{0.48\columnwidth}
\centering
(a) Structural collapse
\end{minipage}\hfill
\begin{minipage}[t]{0.48\columnwidth}
\centering
(b) Collapse vs.\ trade-off balance
\end{minipage}
\caption{
Structural collapse and its impact on the deletion–retention balance on CIFAR-100 (256 designated instances).
(a) We observe structural collapse during unlearning in prior MU studies (Neg \cite{golatkar2020eternal}, Adv \cite{cha2024learning}, and L2UL \cite{cha2024learning}).
We quantify it as the change in affinities between retained-instance embeddings and anchors before and during unlearning.
Larger values indicate greater semantic shift.
(b) We compare average structural collapse (mean over unlearning steps) with the trade-off accuracy defined as the retention–deletion accuracy gap.
Each dot denotes a random trial; Neg lies along the x-axis, with accuracy below 5\%.}
\label{fig:2}
\end{figure}

Furthermore, we empirically observe that structural collapse is closely associated with the deletion–retention trade-off.
The existing works (orange, green, and blue in Figure \ref{fig:2} (a)) \cite{cha2024learning, golatkar2020eternal} progressively exhibit structural collapse as unlearning proceeds.
This collapse arises from drastic embedding shifts, which make it difficult to preserve their semantic connections.
As illustrated in Figure \ref{fig:2} (b), the average structural collapse is negatively correlated with accuracy.
Notably, reducing structural collapse mitigates representational drift, enabling the model to retain meaningful semantic relations throughout unlearning.
This highlights that structural preservation plays a decisive role for MU.

In this work, we propose a novel structure-faithful unlearning framework.
The core idea is to introduce semantic anchors, which prevent retained instances from drifting during unlearning by binding each instance to these anchors (i.e., stakes).
This preserves semantic organization and prevents structural collapse (see Figure~\ref{fig:1}).
Specifically, the semantic anchors are instantiated from the class attributes, such as texture, shape, or typical context.
These attributes are used to prompt a large language model, which generates human-interpretable descriptions \cite{menon2022visual}.
The resulting descriptions are encoded with a semantic encoder (e.g., CLIP) to form anchors. 
Based on these anchors, we define structure as the affinities between embeddings of retained instances and semantic anchors (as illustrated by the anchor–instance relations in Figure \ref{fig:1}).

Our objective is to ensure that this structure remains preserved throughout unlearning.
To this end, we introduce two complementary constraints. 
The first enforces structure-aware alignment to maintain structural consistency before and after unlearning, thereby ensuring semantic consistency among retained instances.
The second applies structure-aware regularization, which penalizes model updates in proportion to their structural importance.
This prioritizes the preservation of parameters most critical to semantic consistency.
While existing methods neglect structure preservation, our framework explicitly stabilizes this organization and avoids collapse by jointly aligning representations and regularizing updates.

To evaluate the effectiveness of ours, we conduct experiments on three tasks: image classification, face recognition, and image-to-image retrieval.
Across all tasks, our method consistently outperforms existing methods, with the margin widening as the number of instances to unlearn increases.
The results show that retaining semantic structure during unlearning not only mitigates structural collapse but also enhances deletion–retention balance and performance. 
The main contributions of this work are as follows:
\begin{itemize}  
\item We conceptualize structural preservation in instance unlearning and observe that it plays a decisive role in improving the deletion–retention balance.
\item We propose a structure-faithful unlearning framework that leverages semantic anchors to preserve the semantic organization of knowledge.
\item We present two constraints that retain the relational structure between embeddings and anchors and regularize updates to structure-sensitive model parameters.
\item Extensive experimental results show that our method consistently improves both deletion–retention balance and generalization performance across diverse tasks.
\end{itemize}

\section{Related Work}
Machine unlearning (MU) removes the influence of specified data while preserving performance on the remainder \cite{bourtoule2021machine}.
The literature falls into two lines of research: exact unlearning \cite{bourtoule2021machine, graves2021amnesiac, wang2022federated} and approximate unlearning \cite{cha2024learning, golatkar2020eternal, foster2024fast}.
Exact MU \cite{bourtoule2021machine, graves2021amnesiac, wang2022federated} aims to produce a model that behaves as if the deleted samples were never used, typically by retraining on data excluding those samples.
Although this guarantees complete removal, it is computationally expensive and requires access to retained data, which limits its practicality \cite{bourtoule2021machine, cao2015towards}.
In contrast, approximate MU \cite{lee2025esc, kim2024negmerge, spartalis2025lotus} seeks to remove the residual influence of the forget set without full retraining.
This approach, which also includes ours, can be categorized based on various characteristics, as summarized in Table \ref{tab:unlearning_comparison} and discussed below.

\emph{Goal.}
The goal of approximate MU can be divided into two directions: undoing \cite{zhou2025decoupled, mehta2022deep} and misclassification \cite{cha2024learning, wu2020adversarial}. 
Undoing aims to reproduce the model that would result from retraining from scratch without using the forget set.
Misclassification seeks to remove targeted information by enforcing incorrect predictions on designated forget instances.
Note that undoing is inherently limited: the retrained model used as a reference still classifies forget instances and varies across runs \cite{goel2022towards}, making it difficult both to guarantee information removal and to serve as a reliable evaluation baseline.
A key observation motivating our work concerns whether removing undesirable knowledge inevitably disrupts the structure of retained knowledge.
We find that preserving the structure of representations among retained instances is essential for achieving a strong deletion–retention balance (see Figure \ref{fig:2}).
Thus, we remove targeted information while keeping the structure of retained knowledge, called structure-preserving unlearning, and adopt misclassification as the deletion criterion.

\emph{Target Granularity.}
Approximate MU can also be distinguished by its target granularity: class-level \cite{tarun2023fast, yoon2022few, ye2022learning} or instance-level \cite{cha2024learning, spartalis2025lotus, mehta2022deep, kim2022efficient, golatkar2020eternal}.
Class-level unlearning removes all samples of a class while maintaining performance on the remaining classes.
Instance-level unlearning removes specific samples, which can be distributed across multiple classes.
Deletion requests usually target individuals rather than entire classes, making instance-level unlearning both realistic and challenging \cite{unlearning_challenge2023, cha2024learning}.
For these reasons, we focus on instance-level unlearning.

\emph{Data Availability.}
Some studies assume access to the retention dataset during unlearning \cite{spartalis2025lotus, golatkar2020eternal, mehta2022deep}, whereas others prohibit such access \cite{lee2025esc, cha2024learning, kim2024negmerge}.
The latter is more practical, since original training data are often unavailable at the time of deletion due to policy restrictions, storage regulations, or capacity limitations \cite{rosenbaum2010data, zhu2019deep}.
We assume that the retention set is inaccessible, and only the pre-trained model and the requested forget data are available.

\begin{table}[t]
\centering
\caption{Comparison of MU methods. 
SP denotes whether structural preservation is considered, while TG and DA indicate target granularity and data availability, respectively.}
\vspace{-3.mm}
\label{tab:unlearning_comparison}
\scriptsize
\resizebox{\columnwidth}{!}{%
\begin{tabular}{r||*{4}{c}}
\Xhline{0.7pt}
\textbf{Method} & \textbf{Goal} & \textbf{TG} & \textbf{DA} & \textbf{SP} \\
\hline
Selective Forget \cite{golatkar2020eternal} & undo & instance & \cmark & n/a \\ 
UNSIR \cite{tarun2023fast} & undo & class & \cmark & n/a \\
Boundary \cite{chen2023boundary} & undo & class & \xmark & n/a \\ 
L2UL \cite{cha2024learning} & misclassify & instance & \xmark & n/a \\ 
LoTUS \cite{spartalis2025lotus} & undo & instance & \cmark & n/a \\ 
\rowcolor{pastellavender} Ours & misclassify & instance & \xmark & \cmark \\
\Xhline{0.7pt}
\end{tabular}}
\end{table}

\begin{figure*}[t]
\begin{center}
\includegraphics[width=0.9\textwidth]{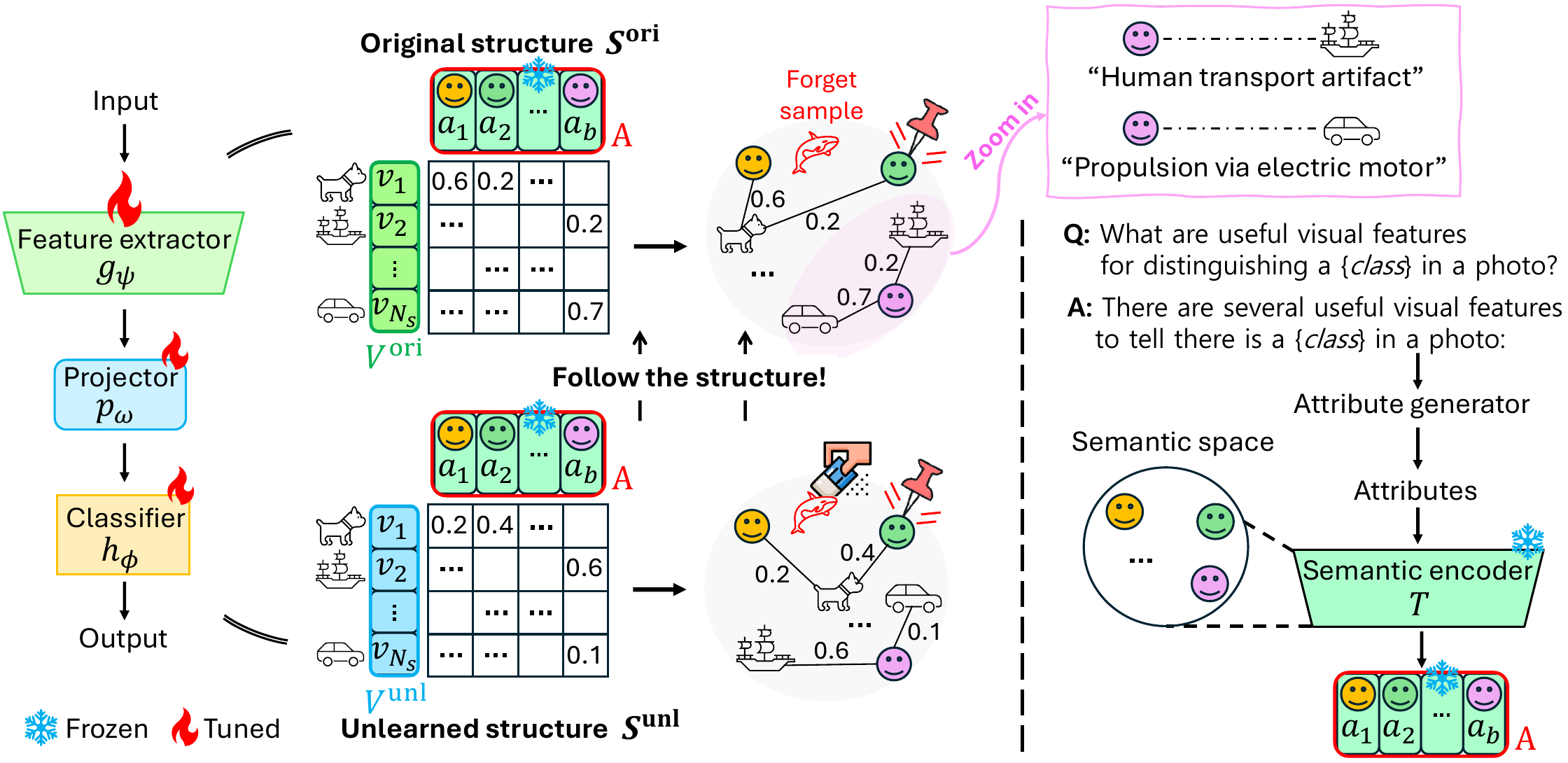}
\vspace{-6.mm}
\end{center}
\caption{
An illustration of the proposed structure-faithful unlearning framework.
The left side illustrates the unlearning process, where we aim to preserve the original structure, $S^{\text{ori}}$, defined by the affinities between visual embeddings $V$ and semantic anchors $A$, by ensuring that these affinities remain consistent in the unlearned structure $S^{\text{unl}}$.
The right side shows the procedure for collecting class-wise anchors: a large language model generates attribute descriptions, which are embedded into anchor vectors via a frozen semantic encoder $T$.}
\label{fig:3}
\end{figure*}

\section{Structure-Faithful Instance Unlearning}
We define the training set as $D_{\text{train}}=\{(x^i,y^i)\}_{i=1}^{N}$, where each input $x^i\in\mathcal{X}$ has its corresponding label $y^i\in\mathcal{Y}$, and $N$ is the total number of instances.
We denote the forget set by $D_f= \{(x_f^i,y_f^i)\}_{i=1}^{N_f}$, which is a subset of $D_{\text{train}}$.
The retention set is defined as $D_r := D_{\text{train}}\setminus D_f$.
We denote the pretrained model as $f_{\theta}^{\text{ori}}(\cdot)=h_{\phi}^{\text{ori}}\big(g_{\psi}^{\text{ori}}(\cdot)\big)$, where $g_{\psi}^{\text{ori}}$ and $h_{\phi}^{\text{ori}}$ are the feature extractor and classifier, respectively.
The model is parameterized by $\theta = \{\psi, \phi\}$, where $\psi$ and $\phi$ represent the sets of parameters of the feature extractor and classifier, respectively.
We aim to obtain an unlearned model $f_{\theta}^{\text{unl}}$ by updating the pretrained model $f_{\theta}^{\text{ori}}$. 
$f_{\theta}^{\text{unl}}$ should eliminate the influence of $D_f$ while preserving performance on $D_r$.
We adopt a realistic regime: during unlearning, we have access only to the pretrained model $f_{\theta}^{\text{ori}}$ and the forget set $D_f$.

As shown in Figure \ref{fig:2} (a), existing MU approaches \cite{cha2024learning, golatkar2020eternal} suffer from progressive structural collapse.
Such collapse indicates that performance degradation arises from a compromised representation structure, underscoring the need for structure-preserving MU.
To this end, we propose a structure-faithful MU framework.
We introduce semantic anchors, which serve as reference points linking retention instances and preserving the overall structure.
To preserve this structure, we propose two constraints: a structure-aware alignment that upholds the distribution of semantic relationships between retained instances and the anchors, thereby maintaining their relative positions, and a regularization that prevents model updates from disrupting them, as detailed in Section \ref{sec:SP}.
The overall framework is shown in Figure \ref{fig:3}.

\subsection{Anchor Generation} 
We aim to remove designated instances while preserving the common semantic relational structure of the retained ones using semantic anchors.
Motivated by finding that interpretable semantic descriptions provide stable reference cues across instances \cite{menon2022visual}, we define an attribute set for each class $c$, $E_c=\{e_{c,j}\}_{j=1}^{M_c}$, by prompting a large language model with a class-aware query \cite{menon2022visual}. 
We concatenate the attributes into a single description $d_c=\operatorname{concat}(e_{c,1},\ldots,e_{c,M_c})$. 
A frozen semantic encoder $T(\cdot)$ maps $d_c$ to the class anchor $a_c$.
We assemble the ($\ell_2$ normalized) anchors into $A=\{a_c\}_{c\in\mathcal{Y}}\in \mathbb{R}^{b \times d}$, where $b$ is the number of classes and $d$ is the embedding dimension.
During unlearning, $A$ remains fixed as a stable and data-frugal reference independent of $D_r$, as illustrated in Figure \ref{fig:3}.

\subsection{Definition of Structure}
We define \textit{structure} as the semantic relations linking anchors and embeddings, inspired by prior studies on relational organization in knowledge \cite{liu2004conceptnet, anderson1983spreading}.
Specifically, we compute affinities that represent the degree of semantic relatedness between anchors $A$ and the embeddings of retention instances produced by the feature extractor.
Since measuring these affinities requires $D_r$, when unavailable, we approximate its embeddings by generating adversarial variants\footnote{Generating $N_{\text{adv}}$ adversarial examples for each $x_f^{i}$, $N_s=N_f\times N_{\text{adv}}$.} from the forget inputs that are pushed toward target-class features \cite{ilyas2019adversarial}. 
This produces a surrogate probe set $D_s=\{(x_s^{\,i},y_s^{\,i})\}_{i=1}^{N_s}$, taking the role of $D_r$ \cite{cha2024learning}.

Let $V^{\text{ori}}=\operatorname{norm}(g_{\psi}^{\text{ori}}(x_s))\in\mathbb{R}^{N_s \times d}$ denote the embeddings from the original pretrained model $f_{\theta}^{\text{ori}}$, where $\operatorname{norm}(\cdot)$ is a normalization operator.
We define the original structure $S^{\text{ori}}$, which represents the state before unlearning, as $V^{\text{ori}} \cdot A^{\top} \in \mathbb{R}^{N_s \times b}$, encoding how the pretrained model organizes knowledge relative to $A$.
Each element of $S^{\text{ori}}$ represents the relevance between an instance and an anchor.
Let $v^{\text{ori}}_i$ and $a_j$ denote the $i$-th and $j$-th rows of $V^{\text{ori}}$ and $A$, respectively, then $S^{\text{ori}}_{i,j} = \langle v^{\text{ori}}_i, a_j \rangle$.
Note that $S^{\text{ori}}$ remains unchanged throughout unlearning.

\noindent\textbf{Unlearned Structure.}
During unlearning, the original structure may shift into an updated relational state due to representation drift.
We quantify this change by measuring the affinities between $A$ and the embeddings obtained during unlearning. 
The resulting relational configuration, which we term the unlearned structure, captures how the underlying knowledge organization evolves through unlearning.
Let $V^{\text{unl}}$ denote the embeddings obtained during unlearning, defined as $\operatorname{norm}\!\big(p_{\omega}(g_{\psi}^{\text{unl}}(x_s))\big)\in\mathbb{R}^{N_s \times d}$, where $p_{\omega}:\mathbb{R}^{d}\!\to\!\mathbb{R}^{d}$ is a learnable projector parameterized by $\omega$ that bridges the feature extractor $g_{\psi}^{\text{unl}}(\cdot)$ and the semantic encoder $T(\cdot)$ to calibrate embeddings for structural preservation.
This projection embeds unlearned features in a semantically coherent space, yielding semantically grounded representations \cite{yu2025language}.
We define the unlearned structure as
\begin{equation}
S^{\text{unl}} = V^{\text{unl}} \cdot A^{\top} \in \mathbb{R}^{N_s \times b}.
\end{equation}
While $S^{\text{ori}}$ captures the relational pattern learned by the pretrained model with respect to the anchors, $S^{\text{unl}}$ quantifies the updated affinities between embeddings and semantic anchors, revealing the structural changes relative to $S^{\text{ori}}$.

\begin{table*}[t]
\caption{
Results of the image classification tasks on CIFAR-10, CIFAR-100, and ImageNet-1K.  
$\mathcal{A}_{\text{test}}$ and $\mathcal{A}_{r}$ represent the accuracies on the test set and the retention set, respectively, evaluated under each value of $k$.
$\mathcal{A}_{f}$ represents the accuracy on the forget set, and we report it as $100 - \mathcal{A}_{f}$, where a higher value indicates more effective unlearning. 
The best performance is highlighted in bold.}
\vspace{-3.mm}
\label{fig:imageclassification}
\resizebox{\textwidth}{!}{%
\begin{tabular}{cl||llll||llll||llll}
\Xhline{1.pt} 
\multirow{2}{*}{} &
  \multirow{2}{*}{\textbf{Method}} &
  \multicolumn{4}{c||}{\textbf{CIFAR-10}} &
  \multicolumn{4}{c||}{\textbf{CIFAR-100}} &
  \multicolumn{4}{c}{\textbf{ImageNet-1K}} \\ \cline{3-14} 
 &
   &
  \multicolumn{1}{c}{$k=16$} &
  \multicolumn{1}{c}{$k=64$} &
  \multicolumn{1}{c}{$k=128$} &
  \multicolumn{1}{c||}{$k=256$} &
    \multicolumn{1}{c}{$k=16$} &
  \multicolumn{1}{c}{$k=64$} &
  \multicolumn{1}{c}{$k=128$} &
  \multicolumn{1}{c||}{$k=256$} &
  \multicolumn{1}{c}{$k=16$} &
  \multicolumn{1}{c}{$k=64$} &
  \multicolumn{1}{c}{$k=128$} &
  \multicolumn{1}{c}{$k=256$} \\ \hline
\multirow{8}{*}{$\mathcal{A}_{\text{test}}$$(\uparrow)$} &
  \textsc{Before} &
  \multicolumn{1}{c}{92.59} &
  \multicolumn{1}{c}{92.59} &
  \multicolumn{1}{c}{92.59} & \multicolumn{1}{c||}{92.59} &
  \multicolumn{1}{c}{77.10} &
  \multicolumn{1}{c}{77.10} &
  \multicolumn{1}{c}{77.10} & \multicolumn{1}{c||}{77.10} &
  \multicolumn{1}{c}{69.79} &
  \multicolumn{1}{c}{69.79} &
  \multicolumn{1}{c}{69.79} & \multicolumn{1}{c}{69.79}
   \\ 
 &
  \textsc{Oracle} &
  \multicolumn{1}{c}{90.21} &
  \multicolumn{1}{c}{91.01} &
  \multicolumn{1}{c}{89.44} & \multicolumn{1}{c||}{38.59} &
  \multicolumn{1}{c}{64.41} &
  \multicolumn{1}{c}{67.06} &
  \multicolumn{1}{c}{66.88} & \multicolumn{1}{c||}{65.31} &
  \multicolumn{1}{c}{63.92} &
  \multicolumn{1}{c}{59.16} &
  \multicolumn{1}{c}{48.18} & \multicolumn{1}{c}{43.35}
   \\ \cline{2-14}
 &
  \textsc{Fisher} &
  \multicolumn{1}{c}{17.10} &
  \multicolumn{1}{c}{14.97} &
  \multicolumn{1}{c}{17.20} & \multicolumn{1}{c||}{15.09} &
  \multicolumn{1}{c}{6.25} &
  \multicolumn{1}{c}{4.69} &
  \multicolumn{1}{c}{2.86} & \multicolumn{1}{c||}{1.18} &
  \multicolumn{1}{c}{53.62} &
  \multicolumn{1}{c}{48.63} &
  \multicolumn{1}{c}{1.44} & \multicolumn{1}{c}{1.70}
   \\ 
 &
  \textsc{Neggrad} &
  \multicolumn{1}{c}{15.87} &
  \multicolumn{1}{c}{9.28} &
  \multicolumn{1}{c}{7.11} & \multicolumn{1}{c||}{6.47} &
  \multicolumn{1}{c}{48.07} &
  \multicolumn{1}{c}{21.11} &
  \multicolumn{1}{c}{10.19} & \multicolumn{1}{c||}{1.71} &
  \multicolumn{1}{c}{53.48} &
  \multicolumn{1}{c}{43.11} &
  \multicolumn{1}{c}{30.74} & \multicolumn{1}{c}{2.09}
   \\ 
 &
  \textsc{Rawp} &
  \multicolumn{1}{c}{63.10} &
  \multicolumn{1}{c}{35.47} &
  \multicolumn{1}{c}{12.91} & \multicolumn{1}{c||}{9.12} &
  \multicolumn{1}{c}{50.99} &
  \multicolumn{1}{c}{14.41} &
  \multicolumn{1}{c}{2.79} & \multicolumn{1}{c||}{1.12} &
  \multicolumn{1}{c}{20.66} &
  \multicolumn{1}{c}{14.28} &
  \multicolumn{1}{c}{4.47} & \multicolumn{1}{c}{2.07} 
   \\ 
 &
  \textsc{Adv} &
  \multicolumn{1}{c}{65.14} &
  \multicolumn{1}{c}{62.23} &
  \multicolumn{1}{c}{49.47} & \multicolumn{1}{c||}{36.69} &
  \multicolumn{1}{c}{63.17} &
  \multicolumn{1}{c}{57.43} &
  \multicolumn{1}{c}{53.89} & \multicolumn{1}{c||}{46.45} &
  \multicolumn{1}{c}{63.44} &
  \multicolumn{1}{c}{57.96} &
  \multicolumn{1}{c}{49.01} & \multicolumn{1}{c}{21.27}
   \\
 &
  \textsc{L2UL} &
  \multicolumn{1}{c}{79.65} &
  \multicolumn{1}{c}{67.08} &
  \multicolumn{1}{c}{50.82} & \multicolumn{1}{c||}{45.44} &
  \multicolumn{1}{c}{63.69} &
  \multicolumn{1}{c}{62.83} &
  \multicolumn{1}{c}{58.44} & \multicolumn{1}{c||}{48.71} &
  \multicolumn{1}{c}{63.92} &
  \multicolumn{1}{c}{59.89} &
  \multicolumn{1}{c}{53.86} & \multicolumn{1}{c}{31.19}
   \\
 &
  \cellcolor{pastellavender} \textsc{Structguard} &
  \multicolumn{1}{c}{\cellcolor{pastellavender}\textbf{84.38}} &
  \multicolumn{1}{c}{\cellcolor{pastellavender}\textbf{69.47}} &
  \multicolumn{1}{c}{\cellcolor{pastellavender}\textbf{58.40}} & \multicolumn{1}{c||}{\cellcolor{pastellavender}\textbf{56.32}} &
  \multicolumn{1}{c}{\cellcolor{pastellavender}\textbf{66.27}} &
  \multicolumn{1}{c}{\cellcolor{pastellavender}\textbf{65.07}} &
  \multicolumn{1}{c}{\cellcolor{pastellavender}\textbf{62.33}} & \multicolumn{1}{c||}{\cellcolor{pastellavender}\textbf{56.91}} &
  \multicolumn{1}{c}{\cellcolor{pastellavender}\textbf{65.53}} &
  \multicolumn{1}{c}{\cellcolor{pastellavender}\textbf{62.57}} &
  \multicolumn{1}{c}{\cellcolor{pastellavender}\textbf{57.59}} & \multicolumn{1}{c}{\cellcolor{pastellavender}\textbf{41.15}}
   \\ \hline\hline
\multirow{8}{*}{$\mathcal{A}_r$$(\uparrow)$} &
  \textsc{Before} &
  \multicolumn{1}{c}{99.60} &
  \multicolumn{1}{c}{99.60} &
  \multicolumn{1}{c}{99.60} & \multicolumn{1}{c||}{99.60} &
  \multicolumn{1}{c}{99.98} &
  \multicolumn{1}{c}{99.98} &
  \multicolumn{1}{c}{99.98} & \multicolumn{1}{c||}{99.98} &
  \multicolumn{1}{c}{79.18} &
  \multicolumn{1}{c}{79.18} &
  \multicolumn{1}{c}{79.18}& \multicolumn{1}{c}{79.18}
   \\ 
 &
  \textsc{Oracle} &
  \multicolumn{1}{c}{98.74} &
  \multicolumn{1}{c}{99.72} &
  \multicolumn{1}{c}{98.97} & \multicolumn{1}{c||}{39.90} &
  \multicolumn{1}{c}{99.96} &
  \multicolumn{1}{c}{96.17} &
  \multicolumn{1}{c}{96.74} & \multicolumn{1}{c||}{96.43} &
  \multicolumn{1}{c}{72.58} &
  \multicolumn{1}{c}{66.37} &
  \multicolumn{1}{c}{52.88} & \multicolumn{1}{c}{46.43}
   \\ \cline{2-14}
 &
  \textsc{Fisher} &
  \multicolumn{1}{c}{17.82} &
  \multicolumn{1}{c}{15.07} &
  \multicolumn{1}{c}{17.43} & \multicolumn{1}{c||}{15.05} &
  \multicolumn{1}{c}{5.31} &
  \multicolumn{1}{c}{4.27} &
  \multicolumn{1}{c}{2.89} & \multicolumn{1}{c||}{1.01} &
  \multicolumn{1}{c}{60.30} &
  \multicolumn{1}{c}{54.87} &
  \multicolumn{1}{c}{1.56} & \multicolumn{1}{c}{1.84}
   \\ 
 &
  \textsc{Neggrad} &
  \multicolumn{1}{c}{15.79} &
  \multicolumn{1}{c}{9.22} &
  \multicolumn{1}{c}{7.11} & \multicolumn{1}{c||}{6.34} &
  \multicolumn{1}{c}{66.97} &
  \multicolumn{1}{c}{26.20} &
  \multicolumn{1}{c}{11.64} & \multicolumn{1}{c||}{1.70} &
  \multicolumn{1}{c}{60.07} &
  \multicolumn{1}{c}{48.05} &
  \multicolumn{1}{c}{34.11} & \multicolumn{1}{c}{2.13}
   \\ 
 &
  \textsc{Rawp} &
  \multicolumn{1}{c}{67.15} &
  \multicolumn{1}{c}{37.32} &
  \multicolumn{1}{c}{13.28} & \multicolumn{1}{c||}{9.12} &
  \multicolumn{1}{c}{72.44} &
  \multicolumn{1}{c}{17.39} &
  \multicolumn{1}{c}{2.82} & \multicolumn{1}{c||}{1.12} &
  \multicolumn{1}{c}{23.48} &
  \multicolumn{1}{c}{16.45} &
  \multicolumn{1}{c}{5.10} & \multicolumn{1}{c}{2.38}
   \\
 &
  \textsc{Adv} &
  \multicolumn{1}{c}{69.70} &
  \multicolumn{1}{c}{66.97} &
  \multicolumn{1}{c}{53.49} & \multicolumn{1}{c||}{39.33} &
  \multicolumn{1}{c}{89.18} &
  \multicolumn{1}{c}{81.07} &
  \multicolumn{1}{c}{76.28} & \multicolumn{1}{c||}{65.67} &
  \multicolumn{1}{c}{72.00} &
  \multicolumn{1}{c}{64.88} &
  \multicolumn{1}{c}{53.90} & \multicolumn{1}{c}{23.43}
   \\ 
 &
  \textsc{L2UL} &
  \multicolumn{1}{c}{85.75} &
  \multicolumn{1}{c}{72.77} &
  \multicolumn{1}{c}{54.51} & \multicolumn{1}{c||}{48.95} &
  \multicolumn{1}{c}{89.81} &
  \multicolumn{1}{c}{89.48} &
  \multicolumn{1}{c}{82.86} &\multicolumn{1}{c||}{67.60} &
  \multicolumn{1}{c}{72.56} &
  \multicolumn{1}{c}{67.28} &
  \multicolumn{1}{c}{59.81} & \multicolumn{1}{c}{35.02}
   \\ 
 &
  \cellcolor{pastellavender} \textsc{Structguard} &
  \multicolumn{1}{c}{\cellcolor{pastellavender}\textbf{91.43}} &
  \multicolumn{1}{c}{\cellcolor{pastellavender}\textbf{76.32}} &
  \multicolumn{1}{c}{\cellcolor{pastellavender}\textbf{64.28}} & \multicolumn{1}{c||}{\cellcolor{pastellavender}\textbf{61.67}} &
  \multicolumn{1}{c}{\cellcolor{pastellavender}\textbf{92.98}} &
  \multicolumn{1}{c}{\cellcolor{pastellavender}\textbf{92.93}} &
  \multicolumn{1}{c}{\cellcolor{pastellavender}\textbf{89.55}} & \multicolumn{1}{c||}{\cellcolor{pastellavender}\textbf{83.30}} &
  \multicolumn{1}{c}{\cellcolor{pastellavender}\textbf{74.62}} &
  \multicolumn{1}{c}{\cellcolor{pastellavender}\textbf{70.56}} &
  \multicolumn{1}{c}{\cellcolor{pastellavender}\textbf{65.01}} & \multicolumn{1}{c}{\cellcolor{pastellavender}\textbf{44.91}}
   \\ \hline\hline
\multirow{8}{*}{$\mathcal{A}_f$$(\uparrow)$} &
  \textsc{Before} &
  \multicolumn{1}{c}{0.00} &
  \multicolumn{1}{c}{0.62} &
  \multicolumn{1}{c}{0.47} & \multicolumn{1}{c||}{0.62} &
  \multicolumn{1}{c}{0.00} &
  \multicolumn{1}{c}{0.00} &
  \multicolumn{1}{c}{0.00} & \multicolumn{1}{c||}{0.00} &
  \multicolumn{1}{c}{0.00} &
  \multicolumn{1}{c}{18.75} &
  \multicolumn{1}{c}{17.97} & \multicolumn{1}{c}{19.54} 
   \\ 
 &
  \textsc{Oracle} &
  \multicolumn{1}{c}{100.00} &
  \multicolumn{1}{c}{100.00} &
  \multicolumn{1}{c}{100.00} & \multicolumn{1}{c||}{99.37} &
  \multicolumn{1}{c}{100.00} &
  \multicolumn{1}{c}{100.00} &
  \multicolumn{1}{c}{100.00} & \multicolumn{1}{c||}{100.00} &
  \multicolumn{1}{c}{100.00} &
  \multicolumn{1}{c}{100.00} &
  \multicolumn{1}{c}{100.00} & \multicolumn{1}{c}{100.00}
   \\ \cline{2-14} 
 &
  \textsc{Fisher} &
  \multicolumn{1}{c}{81.25} &
  \multicolumn{1}{c}{84.38} &
  \multicolumn{1}{c}{78.91} & \multicolumn{1}{c||}{83.59} &
  \multicolumn{1}{c}{93.75} &
  \multicolumn{1}{c}{95.31} &
  \multicolumn{1}{c}{98.44} & \multicolumn{1}{c||}{98.83} &
  \multicolumn{1}{c}{43.75} &
  \multicolumn{1}{c}{29.69} &
  \multicolumn{1}{c}{98.44} & \multicolumn{1}{c}{97.66}
   \\ 
 &
  \textsc{Neggrad} &
  \multicolumn{1}{c}{100.00} &
  \multicolumn{1}{c}{100.00} &
  \multicolumn{1}{c}{100.00} & \multicolumn{1}{c||}{96.17} &
  \multicolumn{1}{c}{100.00} &
  \multicolumn{1}{c}{100.00} &
  \multicolumn{1}{c}{100.00} & \multicolumn{1}{c||}{100.00} &
  \multicolumn{1}{c}{100.00} &
  \multicolumn{1}{c}{100.00} &
  \multicolumn{1}{c}{100.00} & \multicolumn{1}{c}{100.00} \\
 &
  \textsc{Rawp} &
  \multicolumn{1}{c}{100.00} &
  \multicolumn{1}{c}{100.00} &
  \multicolumn{1}{c}{95.94} & \multicolumn{1}{c||}{93.83} &
  \multicolumn{1}{c}{100.00} &
  \multicolumn{1}{c}{100.00} &
  \multicolumn{1}{c}{100.00} & \multicolumn{1}{c||}{100.00} &
  \multicolumn{1}{c}{100.00} &
  \multicolumn{1}{c}{100.00} &
  \multicolumn{1}{c}{100.00} & \multicolumn{1}{c}{100.00}
   \\ 
 &
  \textsc{Adv} &
  \multicolumn{1}{c}{100.00} &
  \multicolumn{1}{c}{100.00} &
  \multicolumn{1}{c}{100.00} & \multicolumn{1}{c||}{100.00} &
  \multicolumn{1}{c}{100.00} &
  \multicolumn{1}{c}{100.00} &
  \multicolumn{1}{c}{100.00} & \multicolumn{1}{c||}{100.00} &
  \multicolumn{1}{c}{100.00} &
  \multicolumn{1}{c}{100.00} &
  \multicolumn{1}{c}{100.00} & \multicolumn{1}{c}{100.00}
   \\
 &
  \textsc{L2UL} &
  \multicolumn{1}{c}{100.00} &
  \multicolumn{1}{c}{100.00} &
  \multicolumn{1}{c}{100.00} & \multicolumn{1}{c||}{100.00} &
  \multicolumn{1}{c}{100.00} &
  \multicolumn{1}{c}{100.00} &
  \multicolumn{1}{c}{100.00} & \multicolumn{1}{c||}{100.00} &
  \multicolumn{1}{c}{100.00} &
  \multicolumn{1}{c}{100.00} &
  \multicolumn{1}{c}{100.00} & \multicolumn{1}{c}{100.00}
   \\
 &
 \cellcolor{pastellavender} \textsc{Structguard} &
  \multicolumn{1}{c}{\cellcolor{pastellavender}100.00} &
  \multicolumn{1}{c}{\cellcolor{pastellavender}100.00} &
  \multicolumn{1}{c}{\cellcolor{pastellavender}100.00} & \multicolumn{1}{c||}{\cellcolor{pastellavender}100.00} &
  \multicolumn{1}{c}{\cellcolor{pastellavender}100.00} &
  \multicolumn{1}{c}{\cellcolor{pastellavender}100.00} &
  \multicolumn{1}{c}{\cellcolor{pastellavender}100.00} & \multicolumn{1}{c||}{\cellcolor{pastellavender}100.00} &
  \multicolumn{1}{c}{\cellcolor{pastellavender}100.00} &
  \multicolumn{1}{c}{\cellcolor{pastellavender}100.00} &
  \multicolumn{1}{c}{\cellcolor{pastellavender}100.00} & \multicolumn{1}{c}{\cellcolor{pastellavender}100.00}
   \\\Xhline{1.pt} 
\end{tabular}}
\end{table*}

\subsection{Structure Preservation}
\label{sec:SP}
To preserve the structure of retained knowledge during unlearning, we ensure that the semantic relations between the retained instances and the anchors remain consistent with those observed before unlearning.
To achieve this, we propose structure-aware alignment.
It enforces consistency between the original and unlearned structures by minimizing their distributional divergence:
\begin{equation}
\mathcal{L}_{\text{align}} = -\frac{1}{b}\sum_{i=1}^{b}\cos\!\big(S^{\text{ori}}_i, S^{\text{unl}}_i\big).
\end{equation}
By maximizing the average cosine similarity between $S^{\text{ori}}$ and $S^{\text{unl}}$, our alignment preserves relative patterns across anchors and instances. 
Empirically, cosine-based alignment yields more stable structural preservation than other measures, such as KL-divergence \cite{kullback1951information} or Wasserstein distance \cite{arjovsky2017wasserstein}, as discussed in Section \ref{sec:analysis}.

While structure-aware alignment provides distributional alignment, it does not account for how model updates affect structural preservation.
Consequently, unregulated model updates can alter structure-critical parameters and weaken semantic consistency.
To complement this, we introduce structure-aware regularization, which restricts model updates to safeguard the semantic information underlying $S^{\text{ori}}$:
\begin{align}
&\mathcal{L}_{\text{reg}} 
= \frac{1}{2}\sum_{i} I_i \cdot \left(\psi^{\text{unl}}_i - \psi^{\text{ori}}_i\right)^2,
\end{align}
where
\begin{align}
&I_i 
= \frac{1}{N_s} \sum_{x_s \in D_s}
\left| \frac{\partial \mathcal{L}_{\text{align}}(x_s)}{\partial \psi^{\text{unl}}_i} \right|.
\end{align}
$I_i$ quantifies the structural importance of the $i$-th parameter in the feature extractor.
This suppresses large updates to crucial parameters while allowing moderate changes in less important ones.
Our regularization is distinguished from a recent approach \cite{cha2024learning} in that it constrains parameters crucial for preserving structure, whereas the prior work targets parameters sensitive to the forget samples.

To ensure structural integrity while balancing retention and deletion, we jointly optimize the structure-preserving constraints (i.e., $\mathcal{L}_{\text{align}}$ and $\mathcal{L}_{\text{reg}}$) with the corresponding objectives.
The retention objective encourages prediction of retained instances via the projector to preserve semantic relationships among representations: $\mathcal{L}_{\text{ret}} = \text{CE}\big(h_{\phi}^{\text{unl}}(p_{\omega}(g_{\psi}^{\text{unl}}(x_s))), y_s\big)$, where CE is the cross-entropy loss.
Conversely, the deletion objective bypasses the projector for effective erasing: $\mathcal{L}_{\text{del}} = -\text{CE}\big(h_{\phi}^{\text{unl}}(g_{\psi}^{\text{unl}}(x_f)), y_f\big)$.
The total loss is the sum of the above losses and is minimized over $\tilde{\theta} = \{\psi, \omega, \phi\}$.

\section{Experiments}
\label{sec:experiments}
\subsection{Setup}
\noindent{\textbf{Scenarios and Datasets.}}
To evaluate our method termed \textsc{Structguard}, we adopted the instance-level unlearning scenario, where misclassification is used as the unlearning criterion \cite{cha2024learning}.
We conducted experiments across three distinct tasks: image classification, face recognition, and image-to-image retrieval.
For image classification, we used CIFAR-10 \cite{krizhevsky2009learning}, CIFAR-100 \cite{krizhevsky2009learning}, and ImageNet-1K \cite{deng2009imagenet}.
For face recognition, we evaluated on Lacuna-10 \cite{golatkar2020eternal}, and for image-to-image retrieval, we utilized CIFAR-10 \cite{krizhevsky2009learning}.
In each dataset, we constructed the forget set $D_f$ by randomly selecting $k$ samples from the training data.
Specifically, following \cite{cha2024learning} for image classification and image-to-image retrieval, we used $k \in \{16, 64, 128, 256\}$, while for face recognition, we adopted $k \in \{3, 7, 9, 12, 64\}$.
The remaining samples were treated as the set $D_r$ \cite{cha2024learning}.

\noindent{\textbf{Compared Methods.}}
We evaluated our method against established unlearning baselines.
\textsc{Before} denotes the pretrained model before unlearning, while \textsc{Oracle} denotes a retrained reference model optimized with positive gradients from $D_r$ and negative gradients from $D_f$.
Our comparison also includes diverse unlearning methods, \textsc{Fisher} \cite{golatkar2020eternal}, \textsc{Neggrad} \cite{golatkar2020eternal}, \textsc{Rawp} \cite{wu2020adversarial}, \textsc{Adv} \cite{cha2024learning}, and \textsc{L2UL} \cite{cha2024learning}.
We report classification accuracy on the test set $D_{\text{test}}$, retention set $D_r$, and forget set $D_f$.
Further details on comparison methods are available in the supplementary material \ref{sec:base}.

\noindent{\textbf{Implementation Details.}}
We followed the training details and adversarial sample generation procedure from \cite{cha2024learning}.
As the feature extractor, we used ResNet-18 \cite{he2016deep} for CIFAR-10 and Lacuna-10, and ResNet-50 \cite{he2016deep} for CIFAR-100 and ImageNet-1K. 
The projector, $p_{\omega}$, comprised two linear layers with a ReLU activation.
For attribute description generation, we adopted GPT-4o to produce class-level attributes \cite{menon2022visual}, with the prompting method and sample descriptions provided in supplementary material \ref{sec:attributegeneration}.
Each class was associated with a single anchor, encoded using a ViT-B/32 semantic encoder \cite{radford2021learning}.
An analysis with alternative semantic encoders is included in the supplementary material \ref{sec:semantictype}.
To stabilize the classifier $h_{\phi}^{\text{unl}}$, we applied the elastic net penalty \cite{zou2005regularization} as classifier regularization, as discussed in Section~\ref{sec:analysis}.

\subsection{Main Results}
\noindent{\textbf{Image Classification.}}
We evaluated \textsc{Structguard} on CIFAR-10, CIFAR-100, and ImageNet-1K for image classification, as presented in Table \ref{fig:imageclassification}.
Most methods achieve $\mathcal{A}_f$ to 100.00\% across all $k$, except \textsc{Fisher}, which keeps incomplete deletion for every $k$, and \textsc{Neggrad} and \textsc{Rawp} on CIFAR-10 at $k=256$.
For CIFAR-10, ours achieves the highest accuracy on both $\mathcal{A}_{\text{test}}$ and $\mathcal{A}_r$ across all $k$. 
Notably, at $k=256$, it surpasses \textsc{Oracle} by 17.73\% on $\mathcal{A}_{\text{test}}$ and 21.77\% on $\mathcal{A}_r$, demonstrating robust knowledge preservation without a retention set. 
\textsc{Fisher}, \textsc{Neggrad}, and \textsc{Rawp} deteriorate rapidly as $k$ increases, since forgetting more samples without a retention set leads to cumulative representation drift. 
We achieve average gains of 13.76\% and 16.05\% over \textsc{Adv} on $\mathcal{A}_{\text{test}}$ and $\mathcal{A}_r$, respectively, and 6.39\% and 7.93\% over \textsc{L2UL}, averaged across different values of $k$. These gains show that structure-preserving improves deletion–retention balance and generalization.

For CIFAR-100, our method consistently outperforms all baselines across all $k$.
As $k$ increases, \textsc{Fisher}, \textsc{Neggrad}, and \textsc{Rawp} exhibit a consistent decline on both $\mathcal{A}_{\text{test}}$ and $\mathcal{A}_r$, and the degradation becomes more pronounced when the model must handle a larger number of classes than CIFAR-10.
Compared with the strongest baseline, \textsc{L2UL}, ours achieves average gains of 4.22\% on $\mathcal{A}_{\text{test}}$ and 7.25\% on $\mathcal{A}_r$ across $k$.
Notably, at $k=256$, our retention accuracy exceeds \textsc{L2UL} by 15.70\%, demonstrating the benefit of structure preservation.
Furthermore, while \textsc{L2UL} suffers a 22.21\% drop on $\mathcal{A}_r$ when $k$ increases from 16 to 256, ours degrades by only 9.68\%, indicating enhanced stability under larger deletion instances.

For ImageNet-1K, across all $k$, our method outperforms all baselines by an average of 21.57\% on $\mathcal{A}_{\text{test}}$ and 25.91\% on $\mathcal{A}_r$.
Notably, compared with \textsc{L2UL}, while its average performance drops by 10.91\% and 12.51\% for $\mathcal{A}_{\text{test}}$ and $\mathcal{A}_r$, respectively, ours shows smaller declines of 8.12\% and 9.90\%.
These results show that ours remains effective on large-scale datasets, underscoring the need to preserve structure for reliable instance unlearning.
We present class-level unlearning results for image classification in the supplementary material \ref{sec:classlevel}.

\begin{table}[t]
\caption{Results of the face recognition task on Lacuna-10.}
\vspace{-3.mm}
\label{fig:facerecognition}
\centering
\resizebox{\columnwidth}{!}{%
\begin{tabular}{cl||*{5}{c}}
\Xhline{1.pt} 
\multirow{2}{*}{} & \multirow[c]{2}{*}{\textbf{Method}} &
  \multicolumn{5}{c}{\textbf{Lacuna-10}} \\
\cline{3-7}
 & & $k=3$ & $k=7$ & $k=9$ & $k=12$ & $k=64$ \\
\hline
\multirow{8}{*}{$\mathcal{A}_{\text{test}}$($\uparrow$)}
& \textsc{Before}    &
  \multicolumn{1}{c}{94.34} &
  \multicolumn{1}{c}{94.34} &
  \multicolumn{1}{c}{94.34} &
  \multicolumn{1}{c}{94.34} &  \multicolumn{1}{c}{94.34} 
   \\
& \textsc{Oracle}    &
  \multicolumn{1}{c}{91.36} &
  \multicolumn{1}{c}{91.47} &
  \multicolumn{1}{c}{92.00} &
  \multicolumn{1}{c}{91.89} &  \multicolumn{1}{c}{86.78} 
   \\ \cline{2-7} 
& \textsc{Fisher}    &
  \multicolumn{1}{c}{23.02} &
  \multicolumn{1}{c}{15.24} &
  \multicolumn{1}{c}{21.42} &
  \multicolumn{1}{c}{19.82} &  \multicolumn{1}{c}{23.24} 
   \\ 
& \textsc{Neggrad}   &
  \multicolumn{1}{c}{61.51} &
  \multicolumn{1}{c}{46.05} &
  \multicolumn{1}{c}{45.09} &
  \multicolumn{1}{c}{44.88} &  \multicolumn{1}{c}{10.02} 
   \\ 
& \textsc{Rawp}      &
  \multicolumn{1}{c}{72.94} &
  \multicolumn{1}{c}{66.84} &
  \multicolumn{1}{c}{59.91} &
  \multicolumn{1}{c}{41.36} &  \multicolumn{1}{c}{10.66} 
   \\ 
& \textsc{Adv}       &
  \multicolumn{1}{c}{72.28} &
  \multicolumn{1}{c}{59.48} &
  \multicolumn{1}{c}{55.62} &
  \multicolumn{1}{c}{51.81} &  \multicolumn{1}{c}{9.27} 
   \\ 
& \textsc{L2UL}      &
  \multicolumn{1}{c}{75.37} &
  \multicolumn{1}{c}{69.08} &
  \multicolumn{1}{c}{64.81} &
  \multicolumn{1}{c}{58.42} &  \multicolumn{1}{c}{12.26} 
   \\ 
& \cellcolor{pastellavender} \textsc{Structguard} &
  \multicolumn{1}{c}{\cellcolor{pastellavender}\textbf{77.29}} &
  \multicolumn{1}{c}{\cellcolor{pastellavender}\textbf{70.36}} &  \multicolumn{1}{c}{\cellcolor{pastellavender}\textbf{70.14}} &
  \multicolumn{1}{c}{\cellcolor{pastellavender}\textbf{64.07}} &
  \multicolumn{1}{c}{\cellcolor{pastellavender}\textbf{27.71}} 
   \\
\hline\hline
\multirow{8}{*}{$\mathcal{A}_{r}$($\uparrow$)}
& \textsc{Before}    &
\multicolumn{1}{c}{100.00} &
  \multicolumn{1}{c}{100.00} &
  \multicolumn{1}{c}{100.00} &
  \multicolumn{1}{c}{100.00} &  \multicolumn{1}{c}{100.00} 
   \\ 
& \textsc{Oracle}    &
\multicolumn{1}{c}{99.19} &
  \multicolumn{1}{c}{99.54} &
  \multicolumn{1}{c}{99.43} &
  \multicolumn{1}{c}{99.52} &  \multicolumn{1}{c}{98.44} 
   \\ \cline{2-7} 
& \textsc{Fisher}    &
\multicolumn{1}{c}{19.84} &
  \multicolumn{1}{c}{13.73} &
  \multicolumn{1}{c}{23.50} &
  \multicolumn{1}{c}{16.97} &  \multicolumn{1}{c}{22.66} 
   \\ 
& \textsc{Neggrad}   &
\multicolumn{1}{c}{68.14} &
  \multicolumn{1}{c}{45.94} &
  \multicolumn{1}{c}{43.74} &
  \multicolumn{1}{c}{40.86} &  \multicolumn{1}{c}{9.57} 
   \\ 
& \textsc{Rawp}      &
 \multicolumn{1}{c}{81.82} &
  \multicolumn{1}{c}{76.60} &
  \multicolumn{1}{c}{66.64} &
  \multicolumn{1}{c}{48.01} &  \multicolumn{1}{c}{11.82} 
   \\ 
& \textsc{Adv}       &
 \multicolumn{1}{c}{80.38} &
  \multicolumn{1}{c}{64.33} &
  \multicolumn{1}{c}{58.49} &
  \multicolumn{1}{c}{55.68} &  \multicolumn{1}{c}{11.31} 
   \\ 
& \textsc{L2UL}      &
\multicolumn{1}{c}{83.60} &
  \multicolumn{1}{c}{76.67} &
  \multicolumn{1}{c}{70.01} &
  \multicolumn{1}{c}{64.83} &  \multicolumn{1}{c}{14.66} 
   \\ 
& \cellcolor{pastellavender} \textsc{Structguard} &
  \multicolumn{1}{c}{\cellcolor{pastellavender}\textbf{84.82}} &
  \multicolumn{1}{c}{\cellcolor{pastellavender}\textbf{77.01}} &  \multicolumn{1}{c}{\cellcolor{pastellavender}\textbf{77.17}} &
  \multicolumn{1}{c}{\cellcolor{pastellavender}\textbf{66.98}} &
  \multicolumn{1}{c}{\cellcolor{pastellavender}\textbf{29.97}} 
   \\
\hline\hline
\multirow{8}{*}{$\mathcal{A}_{f}$($\uparrow$)}
& \textsc{Before}    &
 \multicolumn{1}{c}{0.00} &
  \multicolumn{1}{c}{0.00} &
  \multicolumn{1}{c}{0.00} &
  \multicolumn{1}{c}{0.00} &  \multicolumn{1}{c}{0.00} 
   \\ 
& \textsc{Oracle}    &
  \multicolumn{1}{c}{100.00} &
  \multicolumn{1}{c}{100.00} &  \multicolumn{1}{c}{100.00} &
  \multicolumn{1}{c}{100.00} &
  \multicolumn{1}{c}{100.00} 
   \\ \cline{2-7} 
& \textsc{Fisher}    &
  \multicolumn{1}{c}{100.00} &
  \multicolumn{1}{c}{100.00} &  \multicolumn{1}{c}{66.67} &
  \multicolumn{1}{c}{100.00} &
  \multicolumn{1}{c}{75.00} 
   \\
& \textsc{Neggrad}   &
  \multicolumn{1}{c}{100.00} &
  \multicolumn{1}{c}{100.00} &  \multicolumn{1}{c}{100.00} &
  \multicolumn{1}{c}{100.00} &
  \multicolumn{1}{c}{100.00} 
   \\ 
& \textsc{Rawp}      &
  \multicolumn{1}{c}{100.00} &
  \multicolumn{1}{c}{100.00} &  \multicolumn{1}{c}{100.00} &
  \multicolumn{1}{c}{100.00} &
  \multicolumn{1}{c}{100.00} 
   \\
& \textsc{Adv}       &
  \multicolumn{1}{c}{100.00} &
  \multicolumn{1}{c}{100.00} &  \multicolumn{1}{c}{100.00} &
  \multicolumn{1}{c}{100.00} &
  \multicolumn{1}{c}{100.00} 
   \\
& \textsc{L2UL}      &
  \multicolumn{1}{c}{100.00} &
  \multicolumn{1}{c}{100.00} &  \multicolumn{1}{c}{100.00} &
  \multicolumn{1}{c}{100.00} &
  \multicolumn{1}{c}{100.00} 
   \\ 
& \cellcolor{pastellavender} \textsc{Structguard} &
  \multicolumn{1}{c}{\cellcolor{pastellavender}100.00} &
  \multicolumn{1}{c}{\cellcolor{pastellavender}100.00} &  \multicolumn{1}{c}{\cellcolor{pastellavender}100.00} &
  \multicolumn{1}{c}{\cellcolor{pastellavender}100.00} &
  \multicolumn{1}{c}{\cellcolor{pastellavender}100.00} 
   \\ \Xhline{1.pt} 
\end{tabular}}
\end{table}

\begin{figure}[t]
\begin{center}
\includegraphics[width=\columnwidth]{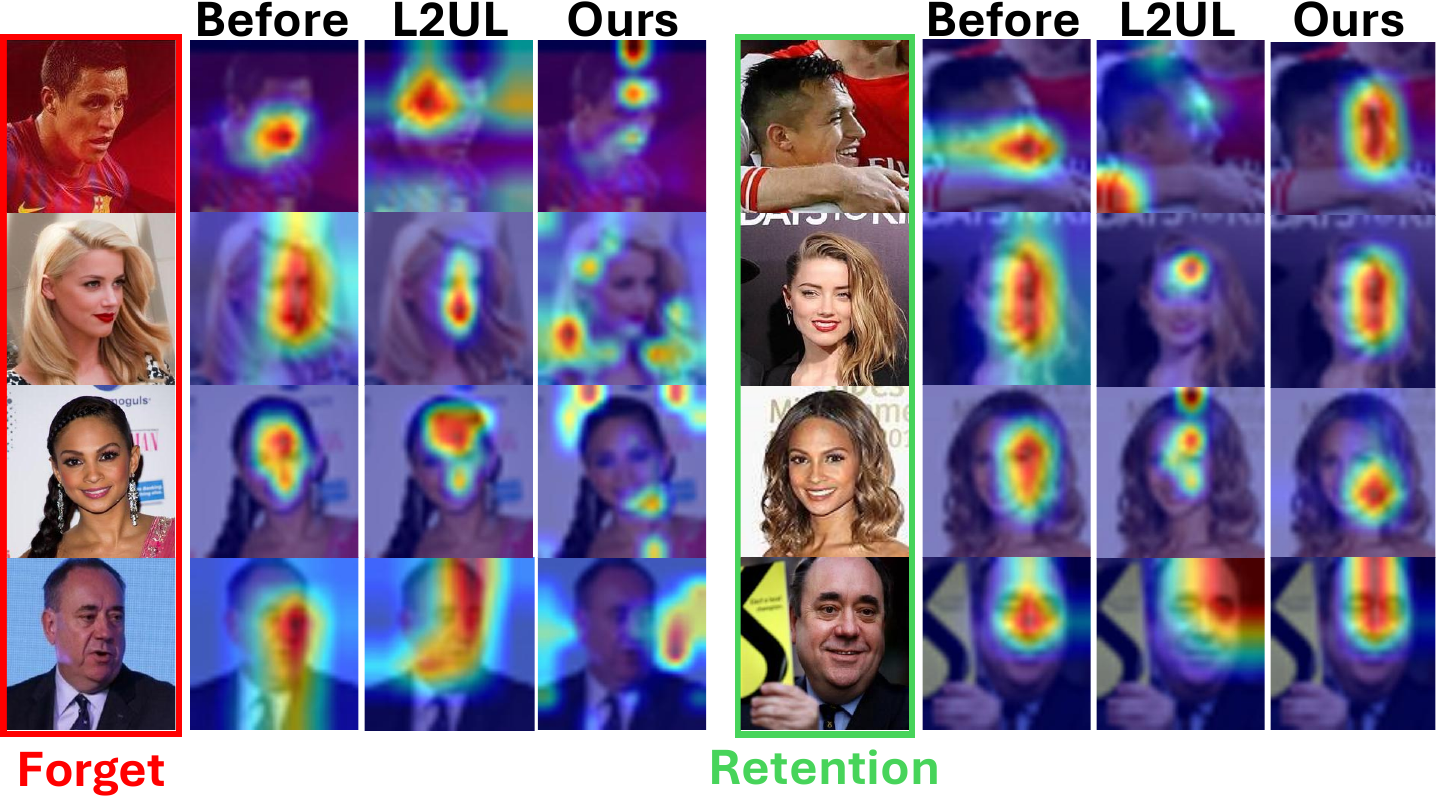}
\end{center}
\vspace{-5.mm}
\caption{Grad-CAM visualizations on Lacuna-10. Red boxes denote forget instances, and green boxes denote retention instances.}
\label{fig:gradcam}
\end{figure}

\begin{figure*}[t]
\includegraphics[width=1.\linewidth]{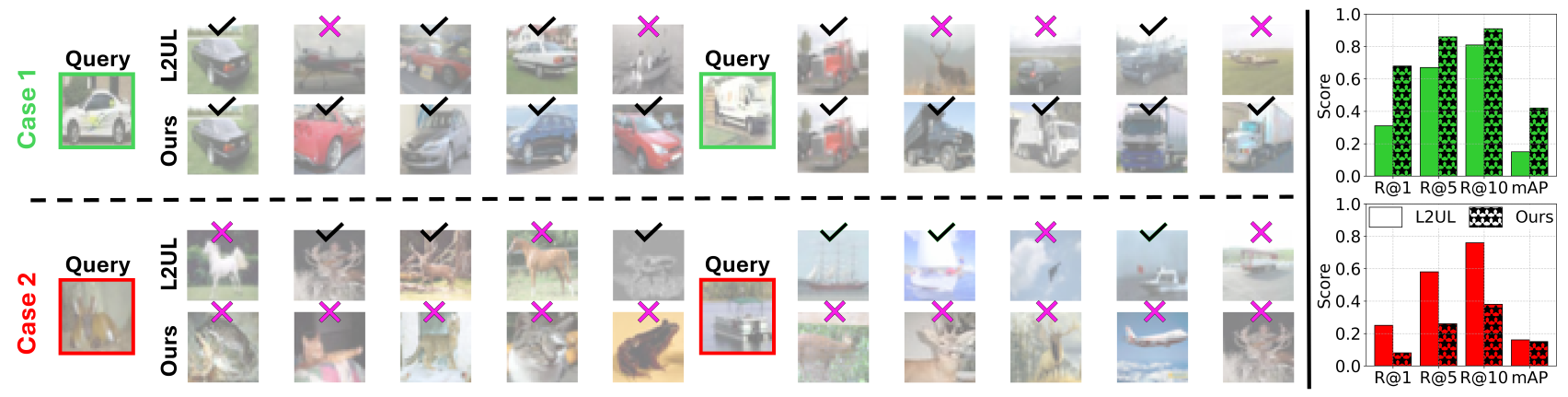}
\vspace{-7.mm}
\caption{Qualitative and quantitative results of the image-to-image retrieval task on CIFAR-10. 
The left side shows top-5 retrieval examples given query images. 
Green-bordered queries correspond to retained samples, while red-bordered queries denote forgotten ones. 
For each query, retrieved images with correct matches are marked with check marks, and incorrect ones with crosses. 
The right bar plots report the quantitative retrieval performance in terms of R@1, R@5, R@10, and mAP, comparing our method with \textsc{L2UL}.}
\label{fig:retrieval}
\end{figure*}

\noindent{\textbf{Face Recognition.}}
To validate the generality of \textsc{Structguard}, we conducted instance unlearning on a face recognition task. 
We used the Lacuna-10 dataset, which is derived from VGG-Faces~\cite{cao2015towards}, following the procedure in \cite{golatkar2020eternal}.
As shown in Table \ref{fig:facerecognition}, except for \textsc{Fisher} at $k=9$ and $k=64$, all methods, including \textsc{Structguard}, successfully remove the designated instances. 
Across all $k$, our method maintains stable recognition performance, achieving gains of 5.92\% and 5.23\% over the strongest baseline, \textsc{L2UL}, on $\mathcal{A}_{\text{test}}$ and $\mathcal{A}_r$, respectively. 
These results show that ours achieves a strong deletion–retention balance and superior generalization, as further evidenced by the Grad-CAM \cite{selvaraju2017grad} results provided in Figure \ref{fig:gradcam}.

\noindent{\textbf{Image-to-Image Retrieval.}}
To extend the evaluation beyond classification, we applied our method to an image-to-image retrieval task as shown in Figure \ref{fig:retrieval}.  
We used the unlearned model under $k=256$ from Table \ref{fig:imageclassification}.
We consider two cases: (i) both the query and retrieved samples belong to the retention set (green), and (ii) the query comes from the forget set while the retrieved samples come from the retention set (red).
The first evaluates knowledge retention quality, while the second examines whether erased instances are isolated in embedding space.

In the first case, our method retrieves same-class samples, while \textsc{L2UL} occasionally returns semantically related but incorrect classes (e.g., \emph{airplane} or \emph{ship} for a \emph{car}), suggesting class confusion.
In the second case, ours retrieves only different-class samples for all forgotten queries (with the supplementary material \ref{sec:misclassification} providing a Streisand effect \cite{golatkar2020eternal} analysis), while \textsc{L2UL} still retrieves same-class samples, implying incomplete deletion.
These results show that ours precisely removes forgotten instances while maintaining feature alignment for retained data.
Quantitatively, Figure~\ref{fig:retrieval} (right) shows that ours achieves higher Recall and mAP for case (i) and lower scores for case (ii), indicating both reliable retention and effective forgetting.

\subsection{Analyses}
\label{sec:analysis}
\begin{figure}[t]
\begin{center}
\includegraphics[width=\columnwidth]{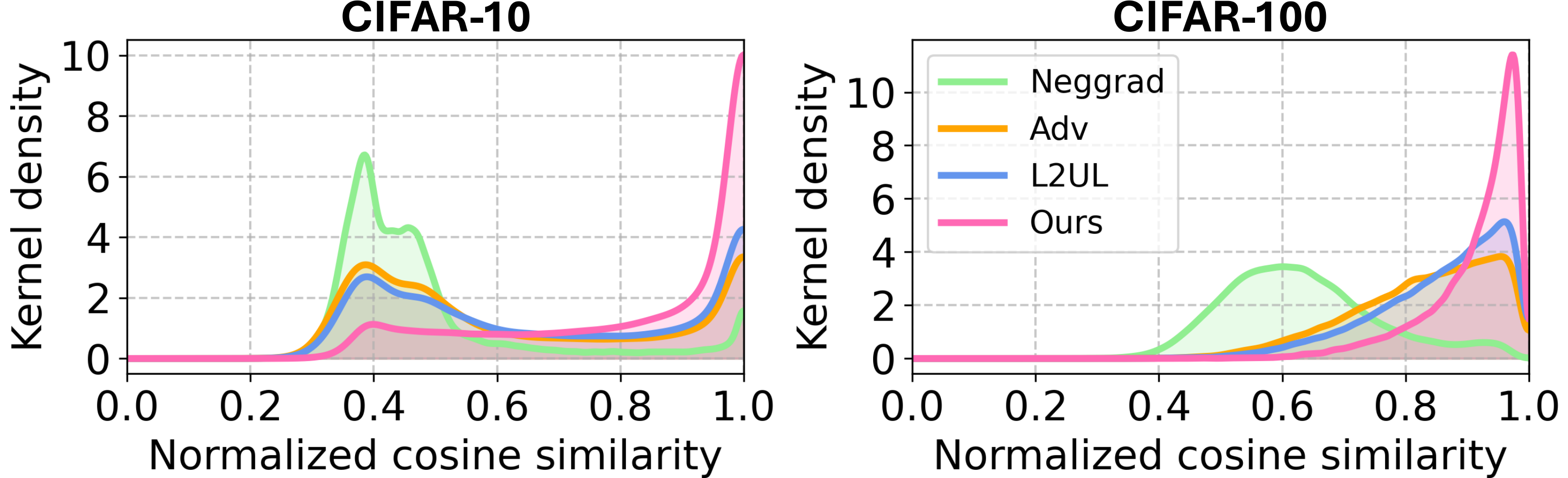}
\end{center}
\vspace{-5.5mm}
\caption{Representation consistency between \textsc{Before} and each unlearning method under $k=256$.}
\label{fig:PDF}
\end{figure}

\noindent{\textbf{Representation Consistency.}}
To assess how unlearning alters the retained representations, we examined their consistency with \textsc{Before}, as shown in Figure~\ref{fig:PDF}. 
Representation consistency is quantified by estimating the kernel density of cosine similarities between the embeddings of identical retained samples from \textsc{Before} and post-unlearning models, where higher similarity indicates smaller deviation from \textsc{Before}.
Across both datasets, \textsc{Neggrad} shows a broad distribution skewed toward low similarity values, showing substantial representation drift. 
Both \textsc{Adv} and \textsc{L2UL} yield a higher proportion of samples with high similarity scores than \textsc{Neggrad}, yet their distributions stay dispersed, with mid-range similarities still appearing on CIFAR-10 and high-similarity values spread widely on CIFAR-100.
Such dispersion indicates inconsistency in the retained representations.
In contrast, ours shows a remarkable result: a sharp peak near 1.0 (unchanged retained representation), indicating that retained features remain well aligned with \textsc{Before}. 
These results show that ours effectively removes target information while retaining the consistency of retained representations with \textsc{Before}.

\noindent{\textbf{Anchor Type.}}
To evaluate the role of semantic anchors in preserving structure, we replaced them with visual prototypes obtained by averaging retention embeddings per class, without linguistic guidance.
As shown in Table \ref{fig:anchortype}, while the visual-anchor variant (Ours-vis) substantially outperforms the non-anchor baseline \textsc{L2UL} on both datasets, semantic anchors (Ours) yield further improvements, increasing $\mathcal{A}_{\text{test}}$ and $\mathcal{A}_r$ by 7.84\% and 7.44\% on CIFAR-10, and by 0.20\% and 2.79\% on ImageNet-1K compared to Ours-vis. 
These results indicate that although anchors are crucial for structural preservation, semantic anchors additionally enhance feature alignment by providing semantically grounded guidance.

\begin{table}[t]
\caption{Comparison of different anchor designs under $k=256$.}
\vspace{-3.mm}
\label{fig:anchortype}
\centering
%\fontsize{5.6pt}{6.8pt}\selectfont  
\resizebox{\columnwidth}{!}{%
\begin{tabular}{r||ccc||ccc}
\Xhline{1.pt} 
\multirow{2}{*}{\textbf{Method}}    & \multicolumn{3}{c||}{\textbf{CIFAR-10}}         & \multicolumn{3}{c}{\textbf{ImageNet-1K}}        \\ \cline{2-7} 
                         & \multicolumn{1}{c}{$\mathcal{A}_{\text{test}}$($\uparrow$)} & $\mathcal{A}_{r}$($\uparrow$) & $\mathcal{A}_{f}$($\uparrow$) & \multicolumn{1}{c}{$\mathcal{A}_{\text{test}}$($\uparrow$)} & $\mathcal{A}_{r}$($\uparrow$) & $\mathcal{A}_{f}$($\uparrow$) \\ \hline
\textsc{L2UL} & \multicolumn{1}{c}{45.44} & \multicolumn{1}{c}{48.95} & 100.00 &\multicolumn{1}{c}{31.19} & \multicolumn{1}{c}{35.02} & 100.00 \\ 
Ours-vis & \multicolumn{1}{c}{48.78} & \multicolumn{1}{c}{54.23} & 100.00 & \multicolumn{1}{c}{40.95} & \multicolumn{1}{c}{42.12} & 100.00 \\ 
\rowcolor{pastellavender} Ours & \multicolumn{1}{c}{\textbf{56.32}} & \multicolumn{1}{c}{\textbf{61.67}} & 100.00  & \multicolumn{1}{c}{\textbf{41.15}} & \multicolumn{1}{c}{\textbf{44.91}} & 100.00 \\ \Xhline{1.pt} 
\end{tabular}}
\end{table}

\begin{figure*}[t]
\includegraphics[
  width=0.82\textwidth,
  height=0.5\textheight,
  keepaspectratio
]{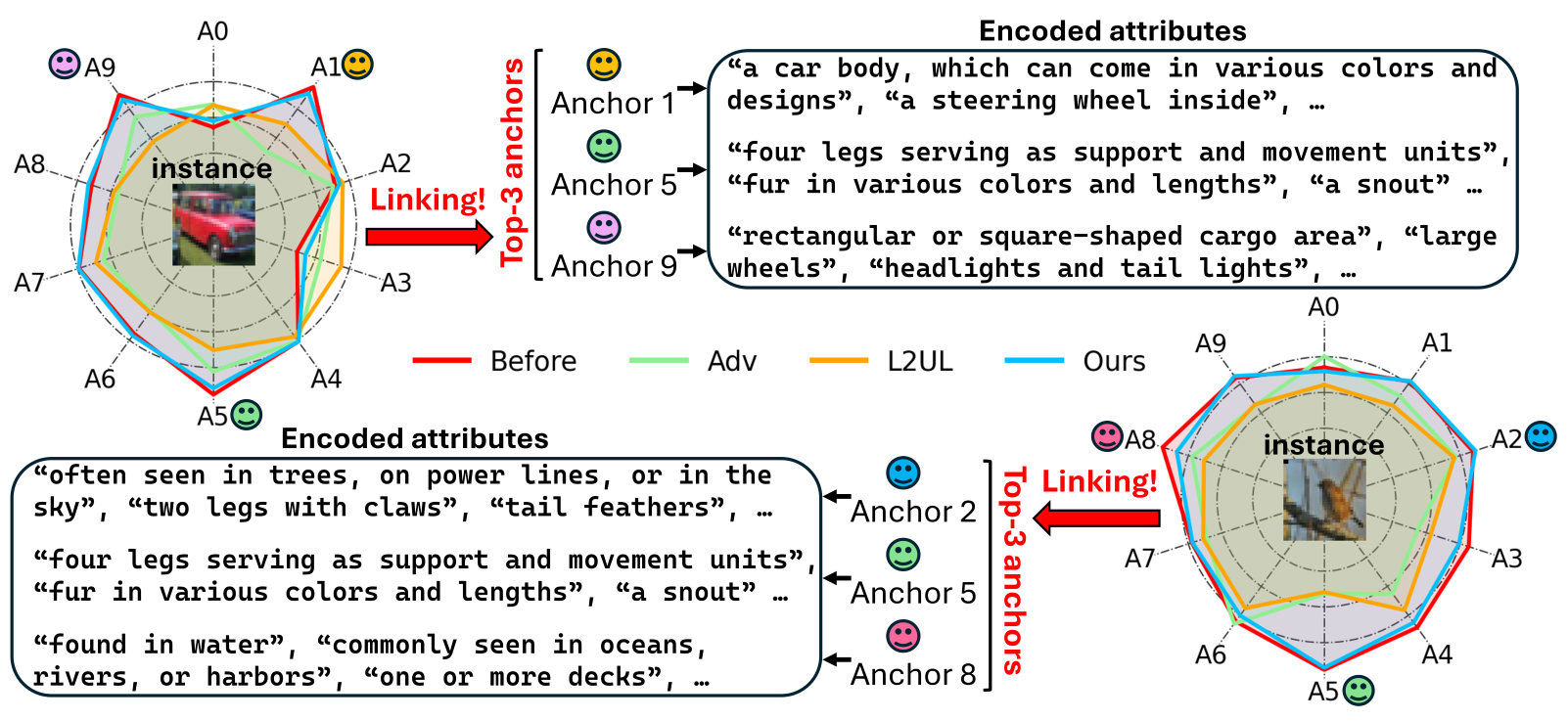}
\centering
\vspace{-3mm}
\caption{Comparison of how the relational structure in \textsc{Before} changes under each unlearning method.
Greater radial extent indicates stronger association with the anchor (A$i$ corresponds to the $i$-th anchor). 
We present the top-3 anchors of the highest affinities along with their corresponding attribute descriptions.}
\label{fig:anchoractivation}
\end{figure*}

\noindent{\textbf{Anchor-Guided Structure Preservation.}}
To investigate whether semantic anchors guide structure preservation, we examined the affinities between randomly selected CIFAR-10 test samples and anchors at $k{=}256$, as shown in Figure~\ref{fig:anchoractivation}.
For both instances, our method yields affinities closely aligned with \textsc{Before}, showing stable semantic associations after unlearning. 
In contrast, \textsc{Adv} and \textsc{L2UL} exhibit noticeable shifts in semantic relations and reduced correspondence to the original anchors, showing structural drift due to the absence of semantic guidance. 
We observe strong connections to the anchors corresponding to their respective classes (Anchor 1 for the left and Anchor 2 for the right).
This reflects that semantic anchors, which encode visual attributes, serve effectively as reference points that support coherent and interpretable structure preservation.

\noindent{\textbf{Ablation Study.}}
We conducted an ablation study to analyze the contribution of each component in \textsc{Structguard}: Structure-aware Alignment (SA), Structure-aware Regularization (SR), and Classifier Regularization (CR), with results in Table~\ref{fig:ablation}.
Removing SA causes the largest performance drop on both datasets, indicating that alignment is the key factor in keeping structural consistency.
On CIFAR-10, the larger performance drop from CR than SR suggests that improvement mainly depends on refining the classifier rather than constraining model updates.
On CIFAR-100, SR becomes more influential than CR as the number of classes increases.
Overall, SA drives structure preservation, while SR and CR enhance stability by regulating model updates and classifier behavior across different class scales.

\begin{table}[t]
\caption{Ablation study on CIFAR-10 and CIFAR-100.}
\vspace{-3mm}
\label{fig:ablation}
\resizebox{\columnwidth}{!}{%
\centering
\begin{tabular}{ccc||ccc||ccc}
\Xhline{1.pt} 
\multirow{2}{*}{\textbf{SA}} & \multirow{2}{*}{\textbf{SR}} & \multirow{2}{*}{\textbf{CR}} & \multicolumn{3}{c||}{\textbf{CIFAR-10}} & \multicolumn{3}{c}{\textbf{CIFAR-100}} \\ \cline{4-9} 
&  &  & \multicolumn{1}{c}{$\mathcal{A}_{\text{test}}$($\uparrow$)} & $\mathcal{A}_{r}$($\uparrow$) & $\mathcal{A}_{f}$($\uparrow$) & \multicolumn{1}{c}{$\mathcal{A}_{\text{test}}$($\uparrow$)} & $\mathcal{A}_{r}$($\uparrow$) & $\mathcal{A}_{f}$($\uparrow$) \\ \hline
& \cmark & \cmark & \multicolumn{1}{c}{37.75} & \multicolumn{1}{c}{42.36} & 100.00 & \multicolumn{1}{c}{51.49} & \multicolumn{1}{c}{80.52} & 100.00 \\ 
\cmark &  & \cmark & \multicolumn{1}{c}{55.01} & \multicolumn{1}{c}{60.67} & 100.00 & \multicolumn{1}{c}{55.86} & \multicolumn{1}{c}{82.33} & 100.00 \\ 
\cmark & \cmark &  & \multicolumn{1}{c}{54.08} & \multicolumn{1}{c}{59.83} & 100.00 & \multicolumn{1}{c}{56.42} & \multicolumn{1}{c}{82.96} & 100.00 \\ 
\rowcolor{pastellavender} \cmark & \cmark & \cmark & \multicolumn{1}{c}{\textbf{56.32}} & \multicolumn{1}{c}{\textbf{61.67}} & 100.00 & \multicolumn{1}{c}{\textbf{56.91}} & \multicolumn{1}{c}{\textbf{83.30}} & 100.00 \\ \Xhline{1.pt} 
\end{tabular}}
\end{table}

\noindent{\textbf{Structural Alignment Loss.}}
To validate the design of the structure-aware alignment loss, we compared cosine similarity (CS) against other measures, such as Mean Squared Error (MSE), Maximum Mean Discrepancy (MMD), Kullback–Leibler (KL) divergence, and Wasserstein distance (WD), as shown in Table~\ref{fig:loss}.
%All losses achieved 100\% forgetting on $\mathcal{A}_f$ across both datasets.
By enforcing directional consistency between features and anchors, CS preserves semantic relationships and maintains structural integrity during unlearning.
On CIFAR-10, CS improves $\mathcal{A}_{\text{test}}$ and $\mathcal{A}_r$ by an average of 9.33\% and 9.54\% over other losses, and it maintains consistently strong performance on CIFAR-100.
These results show that CS provides a simple and effective alignment objective for structure-faithful unlearning.

\begin{table}[t]
\caption{Comparison of different loss functions for structure-aware alignment under $k=256$.}
\vspace{-3mm}
\label{fig:loss}
\centering
\resizebox{\columnwidth}{!}{%
\begin{tabular}{r||ccc||ccc}
\Xhline{1.pt} 
\multirow{2}{*}{\textbf{Loss}}    & \multicolumn{3}{c||}{\textbf{CIFAR-10}}         & \multicolumn{3}{c}{\textbf{CIFAR-100}}        \\ \cline{2-7} 
                         & \multicolumn{1}{c}{$\mathcal{A}_{\text{test}}$($\uparrow$)} & $\mathcal{A}_{r}$($\uparrow$) & $\mathcal{A}_{f}$($\uparrow$) & \multicolumn{1}{c}{$\mathcal{A}_{\text{test}}$($\uparrow$)} & $\mathcal{A}_{r}$($\uparrow$) & $\mathcal{A}_{f}$($\uparrow$) \\ \hline
MSE & \multicolumn{1}{c}{43.80} & \multicolumn{1}{c}{49.67} & 100.00 &\multicolumn{1}{c}{54.50} & \multicolumn{1}{c}{79.72} & 100.00 \\ 
MMD & \multicolumn{1}{c}{43.58} & \multicolumn{1}{c}{47.99} & 100.00 & \multicolumn{1}{c}{55.52} & \multicolumn{1}{c}{80.74} & 100.00 \\ 
KL & \multicolumn{1}{c}{47.41} & \multicolumn{1}{c}{52.13} & 100.00 & \multicolumn{1}{c}{55.46} & \multicolumn{1}{c}{80.35} & 100.00 \\ 
WD & \multicolumn{1}{c}{53.16} & \multicolumn{1}{c}{58.73} & 100.00 & \multicolumn{1}{c}{55.51} & \multicolumn{1}{c}{80.73} & 100.00 \\ 
\rowcolor{pastellavender} CS (Ours) & \multicolumn{1}{c}{\textbf{56.32}} & \multicolumn{1}{c}{\textbf{61.67}} & 100.00 & \multicolumn{1}{c}{\textbf{56.91}} & \multicolumn{1}{c}{\textbf{83.30}} & 100.00 \\ \Xhline{1.pt} 
\end{tabular}}
\end{table}

\section{Conclusion}
We have proposed \textsc{Structguard}, a novel method that removes designated information while preserving the integrity of retained knowledge. 
We introduce anchors constructed from attribute sets generated by a language model and encoded with a semantic encoder.
These anchors act as reference points to preserve representation structure, capturing the semantic relationships between anchors and the embeddings of retained instances.
To maintain structure, we propose two constraints: structure-aware alignment, which aligns the structure before and after unlearning, and structure-aware regularization, which limits changes to parameters critical to structural integrity. 
Experiments on diverse tasks show that our method achieves superior deletion–retention balance and generalization over baselines.

\newpage
\noindent{\textbf{Acknowledgements.}} This work was supported by the Institute of Information \& Communications Technology Planning \& Evaluation (IITP) grant funded by the Korea government (MSIT) [RS-2021-II211341, Artificial Intelligence Graduate School Program (Chung-Ang University)].

{
    \small
    \bibliographystyle{ieeenat_fullname}
    \bibliography{main}
}

\clearpage
\appendix
\begin{center}
{\large \textbf{Supplementary Material}}
\end{center}

\section{Baselines}
\label{sec:base}
We compared our method, \textsc{Structguard}, with representative unlearning approaches~\cite{golatkar2020eternal, wu2020adversarial, cha2024learning} aligned with the misclassification objective and the instance-level unlearning scenario:
\begin{itemize}
    \item \textbf{\textsc{Fisher}}~\cite{golatkar2020eternal}: Estimates the importance of parameters for the forget set and suppresses influential weights according to their effect.
    \item \textbf{\textsc{Neggrad}}~\cite{golatkar2020eternal}: Applies gradient reversal on the forget data to directly drive the model away from the forgotten information.
    \item \textbf{\textsc{Rawp}} (a variant of \textsc{Awp}~\cite{wu2020adversarial}): Repeatedly perturbs model weights using forget data to induce misclassification by destabilizing weights linked to forgotten content.
    \item \textbf{\textsc{L2UL}}~\cite{cha2024learning}: Uses adversarial variants of the forget samples to prevent representation-level forgetting, and constrain parameters sensitive to the forget samples through parameter-level forgetting.
    \item \textbf{\textsc{Adv}} (a variant of \textsc{L2UL}~\cite{cha2024learning}): Performs gradient ascent on the forget set and incorporates adversarial examples generated from it to maintain the retained representations.
\end{itemize}

\section{Class-level Unlearning}
\label{sec:classlevel}
Although our focus is on instance-level unlearning scenarios, we also evaluated whether the proposed method extends to class-level deletion.
Following prior studies \cite{lee2025esc}, we designated 10\% of the entire set of classes as the forget set and conducted experiments on CIFAR-10 and CIFAR-100.
The results are summarized in Table \ref{fig:class}.
All methods successfully remove the designated classes on both datasets, except for \textsc{Fisher} on CIFAR-10.
In particular, \textsc{Fisher} and \textsc{Neggrad} exhibit unstable performance on both datasets, because class-level deletion requires removing all samples belonging to the designated classes, which intensifies structural collapse and hinders retention.
\textsc{Rawp} maintains performance on CIFAR-10 but struggles on CIFAR-100 as the number of forgotten classes grows. 
\textsc{Adv} demonstrates stronger retention than \textsc{Rawp} on CIFAR-100, due to its use of adversarial samples for retention. 
\textsc{L2UL} consistently outperforms all baselines on both datasets, making it the strongest baseline for class-level deletion.
Notably, compared to \textsc{L2UL}, our approach achieves average improvements of 10.45\% on the test-set accuracy ${\mathcal{A}}_{\text{test}}$ and 12.28\% on the retention-set accuracy ${\mathcal{A}}_{r}$, averaged across both datasets.
These noticeable performance gaps reflect the greater structural collapse caused by representation distortion under class-level deletion, underscoring that preserving the relational structure of the retained knowledge becomes even more critical.

\begin{table}[t]
\scriptsize
\caption{Results of the class-level unlearning scenario.}
\vspace{-3mm}
\label{fig:class}
\centering
\resizebox{\columnwidth}{!}{%
\begin{tabular}{cl||*{2}{c}}
\Xhline{0.7pt} 
\multirow{1}{*}{} & \multirow[c]{1}{*}{\textbf{Method}} &
  \multicolumn{1}{c||}{\textbf{CIFAR-10}} &
  \multicolumn{1}{c}{\textbf{CIFAR-100}}\\
\cline{3-4}\hline
\multirow{8}{*}{$\mathcal{A}_{\text{test}}$($\uparrow$)}
& \textsc{Before}    &
  \multicolumn{1}{c||}{92.59} &
  \multicolumn{1}{c}{77.10} 
   \\ 
& \textsc{Oracle}    &
  \multicolumn{1}{c||}{79.14} &
  \multicolumn{1}{c}{57.44} 
   \\ \cline{2-4} 
& \textsc{Fisher}    &
  \multicolumn{1}{c||}{11.84} &
  \multicolumn{1}{c}{1.92} 
   \\ 
& \textsc{Neggrad}   &
  \multicolumn{1}{c||}{13.77} &
  \multicolumn{1}{c}{20.82} 
   \\
& \textsc{Rawp}      &
  \multicolumn{1}{c||}{59.31} &
  \multicolumn{1}{c}{28.14} 
   \\
& \textsc{Adv}       &
  \multicolumn{1}{c||}{62.20} &
  \multicolumn{1}{c}{42.53} 
   \\ 
& \textsc{L2UL}      &
  \multicolumn{1}{c||}{64.56} &
  \multicolumn{1}{c}{43.38} 
   \\ 
& \cellcolor{pastellavender} \textsc{Structguard} &
  \multicolumn{1}{c||}{\cellcolor{pastellavender}\textbf{78.20}} &
  \multicolumn{1}{c}{\cellcolor{pastellavender}\textbf{50.65}} 
   \\ \hline\hline
\multirow{8}{*}{$\mathcal{A}_{r}$($\uparrow$)}
& \textsc{Before}    &
\multicolumn{1}{c||}{99.63} &
  \multicolumn{1}{c}{99.98} 
   \\
& \textsc{Oracle}    &
\multicolumn{1}{c||}{93.06} &
  \multicolumn{1}{c}{87.66} 
   \\ \cline{2-4} 
& \textsc{Fisher}    &
\multicolumn{1}{c||}{12.62} &
  \multicolumn{1}{c}{1.03} 
   \\ 
& \textsc{Neggrad}   &
\multicolumn{1}{c||}{14.94} &
  \multicolumn{1}{c}{26.60} 
   \\ 
& \textsc{Rawp}      &
 \multicolumn{1}{c||}{69.14} &
  \multicolumn{1}{c}{39.98} 
   \\ 
& \textsc{Adv}       &
 \multicolumn{1}{c||}{74.22} &
  \multicolumn{1}{c}{63.38} 
   \\ 
& \textsc{L2UL}      &
\multicolumn{1}{c||}{75.79} &
  \multicolumn{1}{c}{64.32} 
   \\ 
& \cellcolor{pastellavender} \textsc{Structguard} &
  \multicolumn{1}{c||}{\cellcolor{pastellavender}\textbf{93.48}} &
  \multicolumn{1}{c}{\cellcolor{pastellavender}\textbf{71.19}}
   \\\hline\hline
\multirow{8}{*}{$\mathcal{A}_{f}$($\uparrow$)}
& \textsc{Before}    &
 \multicolumn{1}{c||}{0.00} &
  \multicolumn{1}{c}{0.00} 
   \\ 
& \textsc{Oracle}    &
  \multicolumn{1}{c||}{100.00} &
  \multicolumn{1}{c}{100.00} 
   \\ \cline{2-4} 
& \textsc{Fisher}    &
  \multicolumn{1}{c||}{93.68} &
  \multicolumn{1}{c}{100.00} 
   \\ 
& \textsc{Neggrad}   &
  \multicolumn{1}{c||}{100.00} &
  \multicolumn{1}{c}{100.00}  
   \\ 
& \textsc{Rawp}      &
  \multicolumn{1}{c||}{100.00} &
  \multicolumn{1}{c}{100.00} 
   \\
& \textsc{Adv}       &
  \multicolumn{1}{c||}{100.00} &
  \multicolumn{1}{c}{100.00}  
   \\ 
& \textsc{L2UL}      &
  \multicolumn{1}{c||}{100.00} &
  \multicolumn{1}{c}{100.00}  
   \\ 
& \cellcolor{pastellavender} \textsc{Structguard} &
  \multicolumn{1}{c||}{\cellcolor{pastellavender}100.00} &
  \multicolumn{1}{c}{\cellcolor{pastellavender}100.00} 
   \\ \Xhline{0.7pt} 
\end{tabular}}
\end{table}

\begin{table}[t]
\caption{Comparison of different semantic encoders under $k=256$.}
\vspace{-3mm}
\label{fig:semanticencoder}
\centering
\resizebox{\columnwidth}{!}{%
\begin{tabular}{r||ccc||ccc}
\Xhline{1.pt} 
\multirow{2}{*}{\textbf{Encoder}}    & \multicolumn{3}{c||}{\textbf{CIFAR-10}}         & \multicolumn{3}{c}{\textbf{CIFAR-100}}        \\ \cline{2-7} 
                         & \multicolumn{1}{c}{$\mathcal{A}_{\text{test}}$($\uparrow$)} & $\mathcal{A}_{r}$($\uparrow$) & $\mathcal{A}_{f}$($\uparrow$) & \multicolumn{1}{c}{$\mathcal{A}_{\text{test}}$($\uparrow$)} & $\mathcal{A}_{r}$($\uparrow$) & $\mathcal{A}_{f}$($\uparrow$) \\ \hline
%\textsc{L2UL} & \multicolumn{1}{c|}{45.44} & \multicolumn{1}{c|}{48.95} & 100.00 &\multicolumn{1}{c|}{48.71} & \multicolumn{1}{c|}{67.60} & 100.00 \\ \hline
SBERT & \multicolumn{1}{c}{47.22} & \multicolumn{1}{c}{52.64} & 100.00 &\multicolumn{1}{c}{56.66} & \multicolumn{1}{c}{82.63} & 100.00 \\ 
SigLIP & \multicolumn{1}{c}{53.99} & \multicolumn{1}{c}{59.03} & 100.00 & \multicolumn{1}{c}{56.90} & \multicolumn{1}{c}{82.75} & 100.00 \\ 
\rowcolor{pastellavender} CLIP (ours) & \multicolumn{1}{c}{\textbf{56.32}} & \multicolumn{1}{c}{\textbf{61.67}} & 100.00  & \multicolumn{1}{c}{\textbf{56.91}} & \multicolumn{1}{c}{\textbf{83.30}} & 100.00 \\ \Xhline{1.pt} 
\end{tabular}}
\end{table}

\section{Semantic Encoder}
\label{sec:semantictype}
We conducted an analysis to examine how the choice of a semantic encoder influences unlearning.
We selected two representative alternatives to CLIP: Sentence-BERT (SBERT) \cite{reimers2019sentence} and SigLIP \cite{zhai2023sigmoid}.
SBERT is a language-only model that generates sentence embeddings, allowing us to test whether anchors derived solely from linguistic semantics can still guide structure preservation.
SigLIP is a multi-modal encoder trained with a sigmoid-based contrastive objective, serving as a counterpart to CLIP.

As shown in Table \ref{fig:semanticencoder}, SBERT shows the weakest performance among the alternatives, as it lacks alignment with visual information.
Furthermore, SigLIP achieves higher performance than SBERT on both datasets and reaches performance close to CLIP on CIFAR-100. 
Because SigLIP is trained on paired image–text data, its anchors are more closely aligned with the visual representations, which facilitates structure preservation, especially when the classes exhibit rich and diverse semantics. 
Among the semantic encoders, CLIP provides the strongest performance within our framework. 
This is because the global contrastive training objective used in CLIP produces representations that align more naturally with the anchor–instance relational structure.
Notably, across all semantic encoders, ours maintains a favorable deletion–retention trade-off and generalizes well, consistently outperforming all the alternatives; the baseline results are reported in Table 2 in the main paper.

\begin{figure}[t]
\begin{center}
\includegraphics[width=\columnwidth]{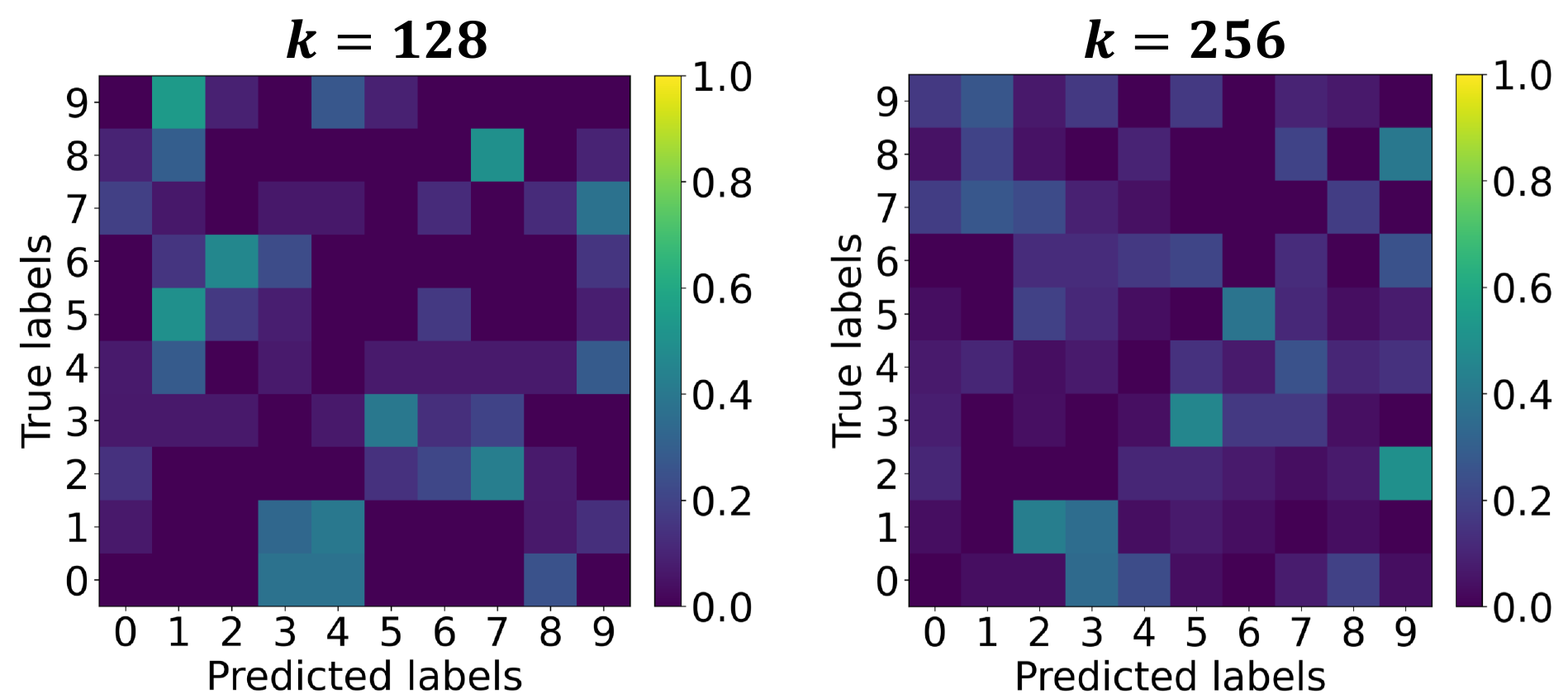}
%\vspace{-10.mm}
\end{center}
\vspace{-6mm}
\caption{Confusion matrices on CIFAR-10 under $k=128$ and $k=256$. 
Each matrix shows post-unlearning predictions on $D_f$, with color brightness proportional to prediction frequency for each label pair.}
\label{fig:missclassification}
\end{figure}

\section{Misclassification}
\label{sec:misclassification}
To test whether our method induces the Streisand effect \cite{golatkar2020eternal}, referring to the unintended exposure of information that should have been forgotten, we examined whether forgotten samples exhibit specific prediction tendencies, as shown in Figure~\ref{fig:missclassification}.
We visualized the prediction tendencies of the unlearned model on the forget set $D_f$ using confusion matrices at $k=128$ and $k=256$ to capture how its outputs deviate from the true labels.
In both confusion matrices, no consistent pattern emerges; the predictions do not form any meaningful arrangement, and this becomes even less organized at $k=256$, indicating no usable regularity that could expose forgotten information.
These results show that our method prevents information leakage of forgotten instances and avoids predictable misclassification, thereby strengthening privacy.

\section{Attribute Generation}
\label{sec:attributegeneration}
Following the previous work \cite{menon2022visual}, we employed prompts in a question and an answer format, as shown in Table \ref{tab:prompt_template}.
These prompts were crafted to obtain concise and distinctive attribute descriptions for each category, which served as the basis for our semantic anchors.
For each dataset, we queried GPT-4o using the question and answer prompts to obtain class-wise attribute descriptions, which are summarized in Table \ref{tab:attribute_examples}.
Each prompt instructed GPT-4o to describe the visual or contextual traits of a given category, including elements such as shape, texture, or common environment.
Prior work \cite{menon2022visual} reported that list formatting becomes more reliable when one or two examples are provided, and we included two short examples in each prompt to encourage structured and coherent outputs.
The resulting class-wise attributes (shown in Table \ref{tab:attribute_examples}) were then encoded with the semantic encoder to construct class-level anchors.
This process produces attribute representations that are semantically coherent and consistently organized, enabling anchor-based structure preservation across datasets.

\begin{table}[t] 
\centering 
\caption{Question–answer prompts used for generating attributes.}
\vspace{-3mm}
\renewcommand{\arraystretch}{1.25} \setlength{\tabcolsep}{10pt} 
\resizebox{\columnwidth}{!}{%
\begin{tabular}{>{\centering\arraybackslash}m{0.22\linewidth} || >{\RaggedRight\arraybackslash}m{0.64\linewidth}} \Xhline{1.pt} 
\textbf{Dataset} & \textbf{Prompts} \\ \hline
\multirow{4}{*}{\makecell[c]{CIFAR-10 \\ CIFAR-100 \\ ImageNet-1K}} & \textbf{Q}: What are useful features for distinguishing a \{\textit{category name}\} in a photo?\\ & \textbf{A}: There are several useful visual features to tell there is a \{\textit{category name}\} in a photo: \\ \hline 
\multirow{5}{*}{Lacuna-10} & \textbf{Q}: What are useful features for distinguishing the face of \{\textit{category name}\} in a photo?\\ & \textbf{A}: There are several useful visual features that help identify the face of \{\textit{category name}\} in a photo, especially in celebrity recognition: \\ \Xhline{1.pt} 
\end{tabular}}
\label{tab:prompt_template} 
\end{table}

\begin{figure}[t]
\begin{center}
\includegraphics[width=\columnwidth]{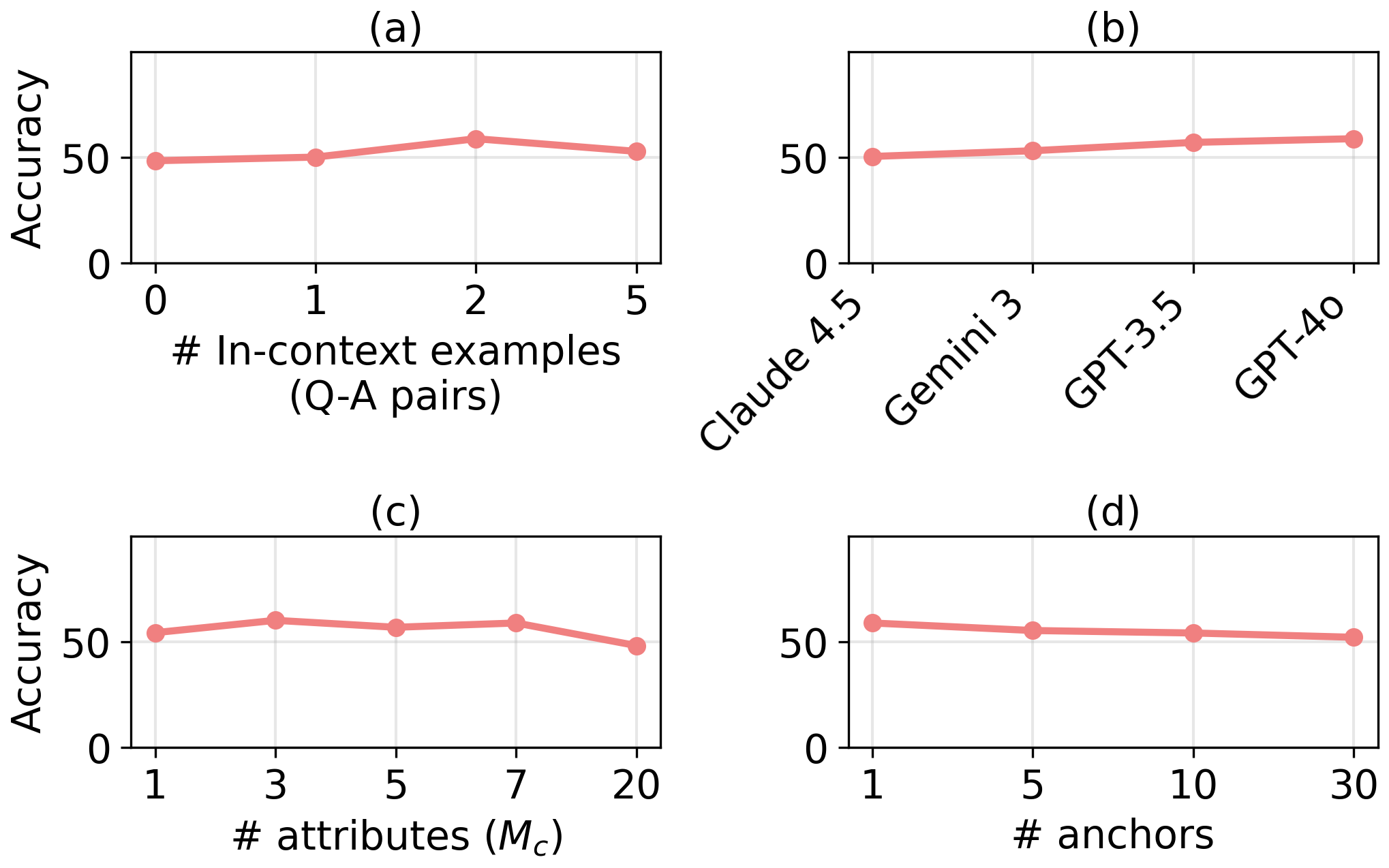}
\end{center}
\vspace{-6.mm}
\caption{Sensitivity analysis on CIFAR-10 ($k=256$).}
\label{fig:AG}
\end{figure}

\section{Sensitivity}
\label{sec:Sensitivity}
Since our anchors are derived from semantic priors provided by an external large language model, assessing the stability of the anchor generation process is important.
To this end, we conducted a sensitivity analysis on (a) the LLM prompt template, (b) the LLM model, (c) the number of attribute descriptions per class, and (d) the number of anchors per class. 
The results are summarized in Figure~\ref{fig:AG}, where accuracy denotes the average of $A_{test}$ and $A_r$.
Across all variations, the proposed method consistently achieves stable performance, demonstrating robustness to the choice of anchor generation components. 
This robustness indicates that semantic anchors provide reliable references for preserving representation structure during unlearning.

\newcolumntype{Y}{>{\RaggedRight\arraybackslash}X} 
\renewcommand\tabularxcolumn[1]{m{#1}}
% ---- table ----
\begin{table*}[t]
\centering
\caption{Examples of attribute descriptions for randomly selected categories.}
\renewcommand{\arraystretch}{1.2}
\setlength{\tabcolsep}{6pt}
\begin{tabularx}{\linewidth}{
>{\centering\arraybackslash}m{0.22\linewidth} ||
>{\centering\arraybackslash}m{0.22\linewidth} ||
Y
}
\Xhline{2\arrayrulewidth}
\textbf{Dataset} & \textbf{Category} & \textbf{Attributes} \\
\Xhline{1\arrayrulewidth}
% CIFAR-10
\multirow{3}{*}[-0.5em]{CIFAR-10} &
airplane &
“wings extending from the sides", “tail fin at the back", “jet engines under the wings or on the tail", “cockpit windows at the front", “landing gear", “airline logos or markings", “propellers if it's a propeller plane" \\
\cline{2-3}
&
cat &
“furry body", “pointed ears", “whiskers", “tail", “round eyes, often green or yellow", “claws", “variety of colors and patterns, including solid, striped, and spotted" \\
\hline
% CIFAR-100
\multirow{3}{*}[-1.em]{CIFAR-100} &
aquarium fish &
“fins and tail", “scales", “gills", “small size", “often found in water or an aquarium setting", “may be seen swimming with other fish", “may have unique patterns or markings on their bodies" \\
\cline{2-3}
&
pickup truck &
“front cab with two or four doors", “a large, open cargo area in the back", “a tailgate at the rear of the cargo area", “a front grille and headlights", “side mirrors", “a license plate", “a truck bed liner or cover (optional)" \\
\hline
% ImageNet-1K
\multirow{3}{*}[-1em]{ImageNet-1K} &
suspension bridge &
“a large structure spanning a body of water or other gap", “typically has two towers or piers supporting the main span", “the main span is suspended by cables or chains", “may have a deck for pedestrians, vehicles, or trains", “may have decorative elements such as lights or flags" \\
\cline{2-3}
&
dust jacket &
“a book cover", “made of paper or cloth", “has a front and a back", “usually has a printed design or image", “may have text on the front and/or back", “may be brightly colored or have a pattern" \\
\hline
% Lacuna-10
\multirow{3}{*}[-1.em]{Lacuna-10} &
Alexis Sánchez &
“intense dark brown eyes and closely shaved eyebrows", “cropped black hair, often styled in a skin fade or taper", “tan skin with high cheekbones and athletic facial lines", “clenched jaw or focused game-time expression", “muscular build, sometimes captured mid-action", “often seen in athletic wear, particularly a football jersey" \\
\cline{2-3}
&
Alesha Dixon &
“almond-shaped dark brown eyes with defined lashes”, “glowing medium-brown skin tone”, “wears bold makeup styles, often with shimmering eye shadow and statement lip colors”, “charismatic smile and dynamic expressions on stage or red carpet”\\
\Xhline{2\arrayrulewidth}
\end{tabularx}
\label{tab:attribute_examples}
\end{table*}


\end{document}